\documentclass[11pt]{article}

\usepackage[final]{acl}

\usepackage{times}
\usepackage{latexsym}

\usepackage[T1]{fontenc}

\usepackage[utf8]{inputenc}

\usepackage{microtype}

\usepackage{inconsolata}

\usepackage{graphicx}
\usepackage{booktabs}
\usepackage{multicol}
\usepackage{amsmath}
\usepackage{amssymb}
\usepackage{multirow} 
\usepackage{tikz} 
\usepackage{pgfplots}
\usepackage{color}
\usepackage{pgfplotstable}
\usepackage{longtable}
\usepackage{tabularx}
\usepackage[dvipsnames,table]{xcolor}
\usepackage{soul}

\usepackage[framemethod=TikZ]{mdframed}
\mdfdefinestyle{hypobox}{
    backgroundcolor=gray!15, 
    linecolor=gray!15,       
    innerleftmargin=10pt,
    innerrightmargin=10pt,
    innertopmargin=10pt,
    innerbottommargin=10pt,
    roundcorner=5pt          
}

\pgfplotsset{compat=1.18}

\usetikzlibrary{patterns}

\newcommand*{\tien}{\textcolor{black}}

\definecolor{gray1}{rgb}{0.93, 0.93, 1.0}
\definecolor{darkgreen}{RGB}{1,50,32}
\definecolor{ForestGreen}{RGB}{34,139,34}
\definecolor{blue1}{HTML}{f1eef6} 
\definecolor{blue2}{HTML}{bdc9e1} 
\definecolor{blue3}{HTML}{74a9cf} 
\definecolor{blue4}{HTML}{2b8cbe} 

\newcommand{\highlightspan}[1]{\colorbox{gray1}{\parbox{0.99\linewidth}{{#1}}}} 
            
\pgfkeys{/pgf/number format/.cd,
    fixed,
    fixed zerofill,
    precision=2,
    set thousands separator={}}
\pgfplotstableread[col sep=space,	
]{pw_ab.tex}\tabPW
\newcommand{\valOfPW}[2]{\pgfplotstablegetelem{#1}{#2}\of\tabPW\pgfmathprintnumber{\pgfplotsretval}}

\pgfplotstableread[col sep=space,columns/*/.style={string type},sci zerofill,precision=2]{ld_ab.tex}\tabLD
\newcommand{\valOfLD}[2]{\pgfplotstablegetelem{#1}{#2}\of\tabLD\pgfmathprintnumber{\pgfplotsretval}}

\pgfplotstableread[col sep=space,columns/*/.style={string type},sci zerofill,precision=2]{prontoQA_ab.tex}\tabProntoQA
\newcommand{\valOfProntoQA}[2]{\pgfplotstablegetelem{#1}{#2}\of\tabProntoQA\pgfmathprintnumber{\pgfplotsretval}}

\pgfplotstableread[col sep=space,columns/*/.style={string type},sci zerofill,precision=2]{prontoQA_ab_olmo.tex}\tabProntoQAOlmo
\newcommand{\valOfProntoQAOlmo}[2]{\pgfplotstablegetelem{#1}{#2}\of\tabProntoQAOlmo\pgfmathprintnumber{\pgfplotsretval}}

\pgfplotstableread[col sep=space,columns/*/.style={string type},]{pw_ab_olmo.tex}\tabPWOlmo
\newcommand{\valOfPWOlmo}[2]{\pgfplotstablegetelem{#1}{#2}\of\tabPWOlmo\pgfmathprintnumber{\pgfplotsretval}}

\pgfplotstableread[col sep=space,columns/*/.style={string type},sci zerofill,precision=2]{ld_ab_olmo.tex}\tabLDOlmo
\newcommand{\valOfLDOlmo}[2]{\pgfplotstablegetelem{#1}{#2}\of\tabLDOlmo\pgfmathprintnumber{\pgfplotsretval}}

\pgfplotstableread[col sep=space,columns/*/.style={string type},sci zerofill,precision=2]{prontoQA_ab_phi4.tex}\tabProntoQAphiF
\newcommand{\valOfProntoQAphiF}[2]{\pgfplotstablegetelem{#1}{#2}\of\tabProntoQAphiF\pgfmathprintnumber{\pgfplotsretval}}

\pgfplotstableread[col sep=space,columns/*/.style={string type},]{pw_ab_phi4.tex}\tabPWphiF
\newcommand{\valOfPWphiF}[2]{\pgfplotstablegetelem{#1}{#2}\of\tabPWphiF\pgfmathprintnumber{\pgfplotsretval}}

\pgfplotstableread[col sep=space,columns/*/.style={string type},sci zerofill,precision=2]{ld_ab_phi4.tex}\tabLDphiF
\newcommand{\valOfLDphiF}[2]{\pgfplotstablegetelem{#1}{#2}\of\tabLDphiF\pgfmathprintnumber{\pgfplotsretval}}

\title{Improving Chain-of-Thought for Logical Reasoning\\via Attention-Aware Intervention}

\author{
 \textbf{Phuong Minh Nguyen}\and
 \textbf{Tien Huu Dang}\and
 \textbf{Naoya Inoue} 
\\
  Japan Advanced Institute of Science and Technology 
\\
\texttt{\{phuongnm,tiendh,naoya-i\}@jaist.ac.jp}
\\
}

\begin{document}
\maketitle




\begin{abstract}
Modern logical reasoning with LLMs primarily relies on employing complex interactive frameworks that decompose the reasoning process into subtasks solved through carefully designed prompts or requiring external resources (\textit{e.g.,} symbolic solvers) to exploit their strong logical structures. 
While interactive approaches introduce additional overhead or depend on external components, which limit their scalability. 
In this work, we introduce a \textit{non-interactive, end-to-end} framework for reasoning tasks, enabling reasoning to emerge within the model itself---improving generalization while preserving analyzability without any external resources.
We show that introducing \textit{structural information into the few-shot prompt activates a subset of attention heads that patterns aligned with logical reasoning operators}. 
Building on this insight, we propose Attention-Aware Intervention (AAI), an \tien{inference-time intervention} method that reweights attention scores across selected heads identified by their logical patterns. 
AAI offers an efficient way to steer the model's reasoning toward leveraging prior knowledge through attention modulation. Extensive experiments show that AAI enhances logical reasoning performance across \tien{diverse} benchmarks, and model architectures, while incurring negligible additional computational overhead. Code is available at \href{https://github.com/phuongnm94/aai_for_logical_reasoning}{\url{https://github.com/phuongnm94/aai_for_logical_reasoning}}.

\end{abstract}

\section{Introduction} 
With the rapid pace of recent LLM developments~\citep{achiam2023gpt,llama3,yang2025qwen3}, logical reasoning has become a crucial capability for many real-world applications that leverage the reasoning abilities of LLMs~\cite{olausson-etal-2023-linc,10.5555/3618408.3618843,pan-etal-2023-logic}. In this task, a question is accompanied by a context consisting of predefined \textit{facts} and \textit{inference rules}. The system is required to perform reasoning over these facts and rules in order to produce the correct answer to the question.

\begin{figure*}[!htbp]
    \centering 
    \includegraphics[width=.99\linewidth, keepaspectratio, 
            trim={0.7cm 0.4cm 0.6cm  0.cm}, page=1, clip=true]{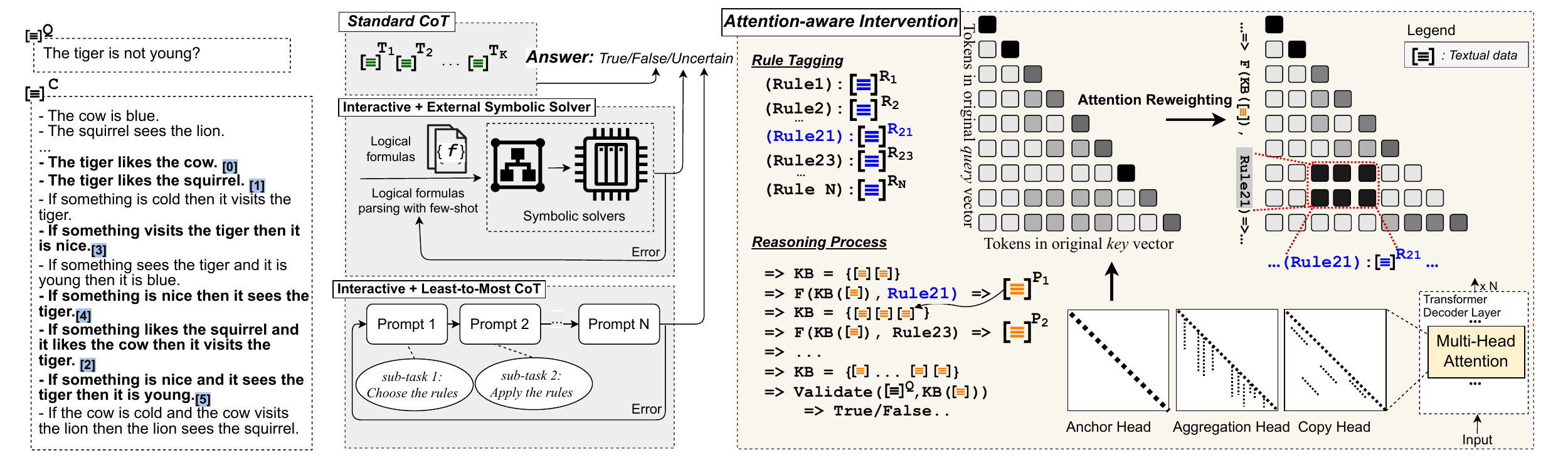}
    \caption{Overview of the comparison between the proposed method and existing approaches.}\label{main_idea}
\end{figure*}

The challenge of logical reasoning tasks lies in the complexity of predefined rules and the large number of reasoning steps that must be processed sequentially. 
In general, the reasoning process can be divided into three sub-steps: assess whether the available information (premises or facts) is sufficient to answer the question; if not, select the appropriate rule(s) to apply; and derive new premises from the given facts and selected rules~\cite{sun-etal-2024-determlr}.
These sub-steps are repeated until the question is answered. 
Fig.~\ref{main_idea} (left) shows a sample from ProofWriter dataset~\citep{tafjord-etal-2021-proofwriter}, where each reasoning step is annotated with numbered tags. Although each step of reasoning \textit{i.e.,} matching conditions and generating new premises is \tien{relatively trivial} for humans, LLMs often make mistakes such as selecting the incorrect rules for the given step or failing to determine when the reasoning process should terminate~\cite{zhang2025cumulative}.

To address this challenge, prior work \tien{has investigated two main approaches.}
The first approach integrates external symbolic solvers (Fig.~\ref{main_idea} middle), delegating the reasoning process to automated theorem provers or programming languages~\cite{ye2023satlm,pan-etal-2023-logic,pmlr-v202-gao23f,xu-etal-2024-symbol}. 
While symbolic solvers can achieve strong performance, they often rely on highly capable LLMs (\textit{e.g.,} ChatGPT) to reliably translate natural language descriptions into formal logical representations (\textit{e.g.,} first-order logic)~\cite{ye2023satlm,pmlr-v202-gao23f}. The second approach constructs frameworks that systematically decompose complex reasoning into sub-tasks such as rule selection, premise inference, and scoring to improve reliability in solving individual steps~\citep{zhang2025cumulative,feng-etal-2024-language,sun-etal-2024-determlr,xu-etal-2024-faithful}. This approach highlights sub-task specialization as a key factor in improving the faithfulness of the reasoning process. However, such methods do not directly evaluate the intrinsic reasoning ability of LLMs, that is, their capacity to \textit{solve the entire problem step by step in a single pass (non-interactive)}. 
\citet{nguyen2025noniterative} introduced Symbolic-Aided CoT prompting, which incorporates a lightweight symbolic structure into the input. This structure abstractly represents fundamental reasoning operators, including rule selection, inference, and knowledge updating. 
However, its efficiency has so far been empirically validated, hindering a better understanding of how and why the injected structural information enhances LLMs' logical reasoning capabilities. In this paper, we \tien{pose a general question:} 
\begin{mdframed}[style=hypobox]
\textit{RQ: How do LLMs encode and utilize structural information in their latent representations for logical reasoning?}
\end{mdframed}

We begin with a preliminary analysis of attention heads in LLM for logical reasoning. \tien{We observe that} injecting symbolic structures into few-shot prompts induces structured activation patterns in certain attention heads, aligning with specific logical reasoning operations. We categorize  attention heads into three functional types. These include \textit{anchor heads}, which are responsible for memorizing information; \textit{aggregation heads}, which integrate newly derived premises and propagate information to the subsequent words of the generation process, and \textit{copy heads}, which replicate newly generated premises when updating the knowledge base or applying matching rules. 

Inspired by recent findings~\citep{sun-etal-2024-determlr,xu-etal-2024-faithful}, which demonstrate that each logical reasoning sub-task \textit{e.g.,} rule selection or new premise generation can be individually inferred by specific few-shot prompts, we introduce Attention-Aware Intervention (AAI), a lightweight and model-agnostic intervention method that enables the model to \textit{focus on focal rules being considered at each inference step while reducing noise from unrelated rules}--as a form of prior knowledge.  
Extensive experiments demonstrate that AAI enhances logical reasoning performance on well-known benchmarks, including ProofWriter, PrOntoQA, Logical Deduction, and FOLIO across model architectures such as Qwen-3, Phi-4, OLMo-2, and sizes ranging from $1.7$B to $32$B while introducing negligible additional computational overhead. 
Moreover, with the Qwen-3 $32$B model, our method achieves competitive performance with state-of-the-art approaches, even compared to those using more powerful closed-source LLMs (\textit{e.g.,} \texttt{gpt-4}) with external interactive modules or symbolic solvers.
\section{Preliminary Experiments\label{sec_preliminary_exp}}
\paragraph{Problem.}
Logical reasoning requires deriving an answer $A$ to a question $Q$ from a set of $N$ contextual rules $\mathcal{R} = \{r_i\}^{N}_i$ where only a relevant subset $\mathcal{R}^* \subset \mathcal{R}$ supports the correct inference graph~\cite{weston2015towards,tafjord-etal-2021-proofwriter}.
 \begin{figure*}[htbp]
    \centering 
    \includegraphics[width=.99\linewidth, keepaspectratio, 
            trim={2.5cm 10.3cm 5cm  3cm}, page=1, clip=true]{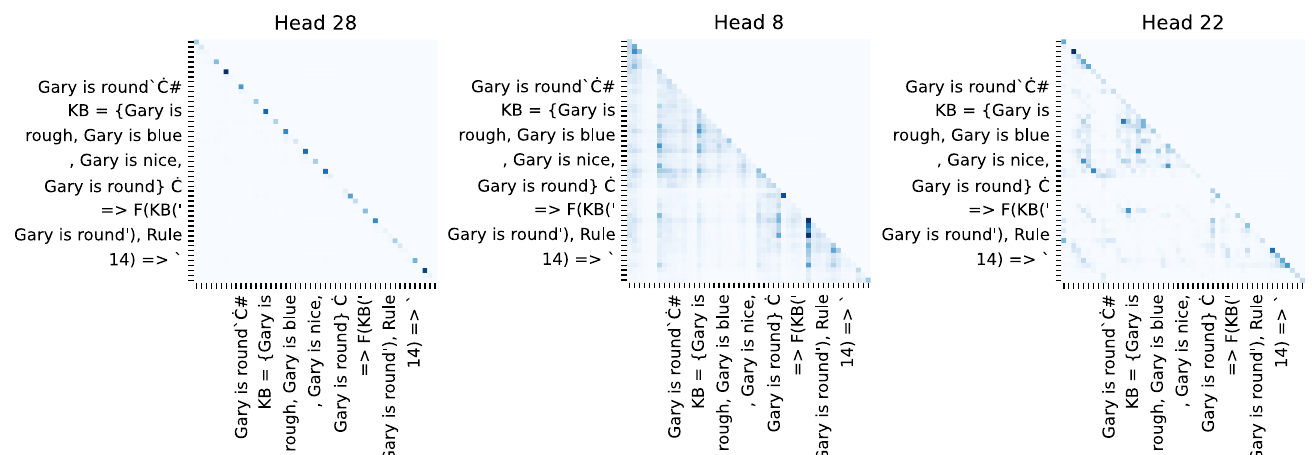}
    \caption{Visualization of attention heads in the \texttt{Qwen3-8B} model on an example (a generated reasoning output) from the ProofWriter dataset. Higher attention scores are represented by stronger colors. For ease of interpretation, every five subword tokens are grouped along both the x- and y-axes in the figure caption. A more detailed (or full-scale) version of this figure is provided in Fig.~\ref{fig_attention_prelim_exp_full}).}\label{fig_attention_prelim_exp}
\end{figure*}
\paragraph{Experimental setting.} In this section, we conduct preliminary experiments to examine the effect of injecting structural information into few-shot CoT prompting on the distribution of attention heads in the Transformer’s multi-head layers. Specifically, we reproduce the Symbolic-Aided CoT method~\cite{nguyen2025noniterative} and analyze the typical patterns of attention scores to explain why Symbolic-Aided CoT achieves significant performance on logical reasoning tasks.

In Symbolic-Aided CoT prompting~\citep{nguyen2025noniterative}, the authors first tag the rules and facts in the logical reasoning problem as variables (\textit{e.g.,} \texttt{Rule1}). A set of symbolic tokens (\textit{e.g.,} \texttt{=>}, \texttt{F}) is then injected into the prompt to represent the structure of the reasoning flow. The inference process is constructed using reasoning operators, including \textit{rule selection}, \textit{new premise inference}, and \textit{knowledge base (KB) updating}. 

The experiments are performed using two-shot Symbolic-Aided CoT on the ProofWriter dataset (Table~\ref{tab_prompting_for_viz} shows a generated reasoning output produced by \texttt{Qwen3-8B} LLM). We visualize the attention matrix values in Fig.~\ref{fig_attention_prelim_exp}.  These values are computed as
$\mathrm{softmax}(qk^{\top} / \sqrt{d})$,
where $q$ and $k$ denote the query and key matrices
in the last transformer decoder layer, and $d$ is their dimensionality~\cite{NIPS2017_3f5ee243}.  The x- and y-axes correspond to the key and query tokens, respectively. 

\begin{table}[!t]
\small
\centering
\caption{\textit{Symbolic-Aided CoT} prompting. The colored text spans were used for attention visualization.
\label{tab_prompting_for_viz} 
}
\resizebox{\linewidth}{!}{%
    \begin{tabular}{|p{1.1\linewidth}|}
    \hline
        ...
        \newline\# (Rule8): Rough things are blue.
        \newline\# (Rule9): Red, quiet things are blue.\newline\# (Rule10): All rough, blue things are nice.  \newline\# (Rule11): If something is quiet and round then it is not nice.\newline\# (Rule12): If Harry is red then Harry is rough.\newline\# (Rule13): If something is nice then it is round.\newline
        \noindent{\fbox{\parbox{0.95\linewidth}{
        \# (Rule14): If something is round then it is red.
        }}}\newline\# (Rule15): Red things are not quiet.\newline\# (Question): Based on the above information, is the following statement true, false, or unknown? Gary is red.\newline\# (Answer): Start from the object and their condition mentioned in the question to collect relevant facts: Gary, is red\newline\# KB = \{\} \newline=> Rule6 = ``Gary is rough''\newline\# KB = \{Gary is rough\} \newline=> F(KB(``Gary is rough''), Rule8) => ``Gary is blue''\newline\# KB = \{Gary is rough, Gary is blue\} \newline=> F(KB(``Gary is blue''), Rule10) => ``Gary is nice''\newline\# KB = \{Gary is rough, Gary is blue, Gary is nice\} \newline=> F(KB(``Gary is nice''), Rule13) => ``\noindent\colorbox{gray1}{Gary is round''}\vspace{-3pt}\newline\noindent\colorbox{gray1}{\# KB = \{Gary is rough, Gary is blue, Gary is nice, Gary is round\}}\vspace{-4pt}
            \newline\noindent\colorbox{gray1}{=> F(KB(``Gary is round''), \noindent{\fbox{Rule14}} ) => }``Gary is red''\newline\# KB = \{Gary is rough, Gary is blue, Gary is nice, Gary is round, Gary is red\} \newline\# valid the question with current infered premies\newline=> Validate(Gary is red, KB(``Gary is red'')) = True. 
        \\
        \hline
    \end{tabular} 
} 
\end{table}
\paragraph{Observations.} 
\textit{Attention structuring.} Firstly, consistent with the observations of~\citet{nguyen2025noniterative}, we found that these logical symbols are represented separately in the hidden semantic space. In the self-attention representations, these symbols receive significantly less attention than informative words such as ``Gary'' or ``round''. This discrepancy causes discontinuities in the self-attention matrix (shown as white or colorless points in Fig.~\ref{fig_attention_prelim_exp}).

\textit{Attention pattern.} Secondly, we found that a subset of attention heads in the self-attention layer reflects the reasoning operators, as shown by the highlighted patterns corresponding to strong word-pair dependencies. In the Transformer decoder’s self-attention layer, the information of a query token vector is aggregated from previous token vectors, with the proportion of contribution represented in the attention matrix (further detail in Appendix~\ref{apd_attn_map_explain}). In Fig.~\ref{fig_attention_prelim_exp} - Head 28, each token primarily attends to itself, without incorporating information from other tokens. The highlighted points mainly correspond to informative words (\textit{e.g.,} ``Gary'', ``round''). We identify this type of head as an \textit{anchor head}, which serves to retain premise-related information. In the second image of Fig.~\ref{fig_attention_prelim_exp} (Head 8), the newly generated premise tends to distribute its information to subsequent inference steps. For example, the premise ``Gary is round'' contributes to the representations of many following tokens, corresponding to the vertical highlights observed in the attention map. We identify this type of head as an \textit{aggregation head}, which supports matching the generated premises with the applicable rules in the logical inference process.
The last image (Head 22) presents another attention pattern characterized by short diagonal alignments, reflecting the copying of an entire information span to another span. For instance, the premise ``Gary is round'' is copied to the knowledge base (KB) variable after it is generated. We identify this type of head as a \textit{copy head}. 


\paragraph{Motivation.} Building on the previous observations, we address the second research question (\textit{RQ2}): \textit{how can we modify the attention matrix to effectively inject prior knowledge into LLMs?} In the Symbolic-Aided CoT framework, each rule is tagged with a symbolic identifier (\texttt{Rule<index>}), requiring the model to retrieve and maintain the corresponding rule information throughout the reasoning process. However, this design can lead to information decay in long reasoning chains, as the identifiers and their associated rule contents often appear in distant spans (\textit{e.g.,} \texttt{Rule14} in Table~\ref{tab_prompting_for_viz}). Furthermore, in copy heads, the copying behavior typically occurs within the same or paraphrased token spans, whereas the original rule content should instead be propagated to the representation of its identifier during reasoning. Similarly, in anchor heads, attention tends to focus on the identifiers themselves rather than capturing the semantics of the underlying rule content.

To address this issue, we propose a mechanism that first categorizes \textit{anchor heads} and \textit{copy heads} and then adaptively reweights them to reinforce the linkage between rule identifiers and their original rule representations. In essence, this process injects structural prior knowledge into the attention mechanism, ensuring that \textit{each rule reference remains grounded in its semantic content during reasoning}.

\section{Methodology\label{sec_method}}
    In this section, we present the details of our proposed method for analyzing attention head patterns to select attention head patterns (\textit{e.g.,} \textit{anchor heads}), and our \textit{Attention-Aware Intervention} mechanism for reweighting their attention scores to inject prior knowledge into LLMs.
    
    Transformer decoder-only architectures are the dominant design for recent LLMs~\citep{radford2019language, llama3, yang2025qwen3}. In these models, semantic features representing the meaning of an input sequence are computed within the multi-head attention layers. Specifically, the self-attention mechanism captures dependencies among word pairs. For generation tasks, a causal mask ($M = M^\mathrm{Causal}$) is applied to prevent tokens from attending to future positions \citep{NIPS2017_3f5ee243}. Given, query and key matrices, $q, k \in \mathbb{R}^{L\times d}$ general formulation of attention is: 
    \begin{align} 
    S &=  \frac{q k ^\top} {\sqrt{d}} \label{eq_dotproduct} \\
    A &= \mathrm{softmax}\left(S + M\right) \label{eq_softmax_a}
    \\
    h^\mathrm{attn}&=A \cdot v 
    \end{align}
    Beyond this standard use, masked matrices have also been explored as a means of injecting prior knowledge into Transformer architectures.
    For example, in neural machine translation, masks have been employed to encode syntactic dependency structures~\citep{bugliarello-okazaki-2020-enhancing,wang-etal-2019-tree} or to model localness~\citep{yang-etal-2018-modeling}. Similarly, in emotion recognition tasks, masks have been utilized to represent intra- and inter-utterance relationships \citep{liu-etal-2022-dialogueein, ijcai2022p628}. However, in the context of in-context learning with LLMs, no training is performed, which makes customizing masks for prior knowledge injection significantly more challenging. Motivated by these observations, we propose an efficient method to inject \textit{reference information} into the decoder-only Transformer architecture by reweighting the masked matrix $M$.
\subsection{Attention Head Analysis}  
    In this step, our objective is to identify attention patterns by analyzing the relative positional relationships between token pairs exhibiting strong dependencies.
    Given an attention score matrix $A \in \mathbb{R}^{L\times L}$ (as defined in Eq.~\ref{eq_softmax_a}), where $L$ denotes the length of the input sequence, each element $(A_{ij})$ represents the dependency score between the $i$-th query token and the $j$-th key token. 
      \begin{center}
        \fontsize{10}{10}\selectfont
        \begin{align}   H_{ij} &=  
            \begin{cases}
              1 & \text{if  $(A_{i,j} > \mathrm{threshold}) \wedge  (i \geq j) $} \\
              0  & \text{otherwise}
            \end{cases} \notag \\
            s^\mathrm{diagonal} &= \frac{\sum_{i,j \mid i,j<L-1}(H_{i,j} \wedge H_{i+1, j+1})} {\sum_{i,j} H_{i,j}} \\
            s^\mathrm{vertical} &= \frac{\sum_{i,j\mid j<L-1}(H_{i,j} \wedge H_{i, j+1})} {\sum_{i,j}H_{i,j}} \\
            s^\mathrm{horizontal} &= \frac{\sum_{i,j\mid i<L-1}(H_{i,j} \wedge H_{i+1, j})} {\sum_{i,j}H_{i,j}}  \\ 
            \textrm{pattern}(A) &= (s^\mathrm{diagonal},s^\mathrm{vertical}, s^\mathrm{horizontal}) \label{eq_head_pattern}
        \end{align}
    \end{center}
    where \textit{threshold} is a hyperparameter used for binarization. The head's pattern is characterized by directional scores: $s^\mathrm{diagonal}$, $s^\mathrm{vertical}$, and $s^\mathrm{horizontal}$, which quantify the relative density of attention edges along the diagonal, vertical, and horizontal directions, respectively. Aggregation heads  exhibit high vertical scores, reflecting strong dependencies between informative tokens. For instance, in Head 8 of Fig.~\ref{fig_attention_prelim_exp}, the token \textit{``round''} receives concentrated attention from subsequent query tokens. In contrast, anchor heads generally display high diagonal but low vertical and horizontal scores, indicating self-focused attention patterns.
\subsection{Attention Head Reweighting} 
    This component is designed to inject human prior knowledge into LLMs through a masked matrix ($M$). By default, this matrix is defined as the causal mask ($M^\mathrm{Causal}$), which prevents information flow from future tokens during generation. In Symbolic-Aided CoT prompting for logical reasoning tasks, the rule identifier tokens (\texttt{Rule<number>}) should be semantically linked with their corresponding rule content.
To enable this, we construct two subsets of token pairs: $\mathcal{D}^\mathrm{Ref}$, which includes pairs connecting a rule identifier to tokens within its own rule span, and $\mathcal{D}^\mathrm{NoRef}$, which includes pairs connecting a rule identifier to tokens' index in other rules. Intuitively, dependencies in $\mathcal{D}^\mathrm{Ref}$ should be encouraged, whereas those in $\mathcal{D}^\mathrm{NoRef}$ should be suppressed. For instance, in Fig.~\ref{main_idea}, \texttt{Rule21} illustrates such relationships within the $\mathcal{D}^\mathrm{Ref}$ set.

Given the scaled dot-product values $S$ computed by Eq.~\ref{eq_dotproduct}, and considering that the range of hidden-state values differs across attention heads, we adaptively compute median-based normalization for each head during the reweighting process. Accordingly, we define two masking matrices, $M^\mathrm{Ref}_{i,j}$ and $M^\mathrm{NoRef}_{i,j}$, to reinforce the representation of rule identifiers throughout the reasoning process. These matrices can be applied independently or combined to integrate both effects, and are used to re-weight selected attention heads in the previous step.
    \begin{center}
        \fontsize{9.6}{9.6}\selectfont
        \begin{align} 
            M^\mathrm{Causal}_{i,j}&=\begin{cases}
              -\infty &\textrm{if}\quad  i < j  \\
              0  & \text{otherwise}
            \end{cases} \\
            M^\mathrm{NoRef}_{i,j}&=\begin{cases}
              -\infty &\textrm{if}\quad  (i, j) \in \mathcal{D}^\mathrm{NoRef} \\
              0  & \text{otherwise}
            \end{cases} \\
            M^\mathrm{Ref}_{i,j}&=\begin{cases}
             c \times \textrm{median}(S) + b &\textrm{if}\,\, (i, j) \in \mathcal{D}^\mathrm{Ref}  \\
              0  & \text{otherwise}
            \end{cases} \label{eq_reweighting} \\
            M^\mathrm{Final} &= M^\mathrm{Ref} + M^\mathrm{NoRef} + M^\mathrm{Causal}
        \end{align}
    \end{center}
   Here, $c$ (reweighting coefficient) and $b$ (reweighting bias) are configuration parameters that control the degree of reweighting (Eq.~\ref{eq_reweighting}). Larger values of $c$ strengthen the connections within the $\mathcal{D}^\mathrm{Ref}$ set in alignment with the original attention matrix values. In contrast, increasing the constant $b$ introduces a uniform bias that does not depend on the original attention values.

\section{Experiments and Result Analysis\label{sec_experiments}}
     In this section, we present the experimental settings, datasets, and evaluation metrics used to assess our proposed method. We then report the main results compared with several strong baselines across multiple logical reasoning benchmarks. Finally, we provide an in-depth analysis and ablation studies to investigate the effectiveness of each component in our approach.
    \subsection{Experimental Setup}
        \paragraph{Datasets.}
        We evaluate our approach on four widely used benchmark datasets for logical reasoning and one mathematical reasoning dataset.
        (1) \textbf{ProofWriter}~\cite{tafjord-etal-2021-proofwriter}: we adopt the subset constructed under the open-world assumption, where each instance can take one of three possible labels--true, false, or unknown. Following the experimental setup of \citet{pan-etal-2023-logic}, we focus on the subset with the highest reasoning complexity (five-hop reasoning), which comprises $600$ examples.
        (2) \textbf{ProntoQA}~\cite{saparov2023language}: similar to ProofWriter, this dataset contains multi-hop logical reasoning problems. Consistent with prior studies~\cite{qi2025large,nguyen2025noniterative}, we use the most challenging subset that requires five-hop reasoning, containing $500$ evaluation samples.
        (3) \textbf{LogicalDeduction}~\cite{srivastava2023beyond}: this dataset evaluates the ability to deduce the correct ordering of entities based on a set of logical constraints. Following the configuration of~\citet{sun-etal-2024-determlr}, we use $300$ samples that include subsets of three, five, and seven objects, where an increased number of objects corresponds to greater reasoning complexity.  
        (4) \textbf{FOLIO}~\cite{han-etal-2024-folio}: this dataset is constructed by domain experts, in which logical rules are compiled using real-world knowledge. Following the experimental settings of previous work~\cite{sun-etal-2024-determlr}, we use the validation subset of this dataset, comprising $204$ examples, for evaluation.
        (5) \textbf{GSM8k}~\citep{gsm8k}: GSM8k is a collection of high-quality, linguistically diverse grade-school math word problems. We evaluate our method on the test subset, consisting of $1319$ examples, to assess our method's generalization capability.
 
        \paragraph{Evaluation metric.}
        We employ accuracy as the evaluation metric, which is a standard measure in prior work~\citep{sun-etal-2024-determlr,qi2025large,nguyen2025noniterative}. This choice enables direct and consistent comparison of our results with existing methods on the same benchmarks.

        \paragraph{Experimental settings.}  
        
        We conducted our experiments primarily on open-source LLMs, including \texttt{Qwen-3}~\cite{yang2025qwen3}, OLMo~\cite{groeneveld2024olmoacceleratingsciencelanguage}, and Phi-4~\cite{abdin2024phi4technicalreport}. 
        The primary objective of this study is to evaluate the effectiveness of the proposed \textbf{A}ttention-\textbf{A}ware \textbf{I}ntervention (\textbf{AAI})  in comparison with the baseline \textit{Symbolic-Aided CoT} \citep{nguyen2025noniterative} and standard CoT prompting. To further validate the generalizability of the AAI mechanism, we introduce a new structured reasoning prompt, termed \textit{Compact Symbolic-Aided CoT}, which employs symbolic representations to provide a minimal structure for the reasoning process. The prompting templates used in both the original \textit{Symbolic-Aided CoT} and the proposed \textit{Compact} version are provided in Appendix~\ref{apd_prompting}. 
        
        We design two experimental settings:
        (1) Main AAI setting -- we perform attention head analysis to select \textit{anchor heads} and \textit{copy heads} with $s^\mathrm{diagonal} > 0.3$, then apply the Attention Head Reweighting mechanism to these heads;
        (2) AAI$^\mathrm{agg.}$ setting -- we apply the same reweighting mechanism to \textit{aggregation heads}, selected under the conditions $s^\mathrm{vertical} > 0.6$ and $s^\mathrm{horizontal}, s^\mathrm{diagonal} < 0.3$. 
        In addition, we conducted extensive experiments to analyze the sensitivity of model performance to hyperparameters, including the binarization, diagonal score thresholds, reweighting coefficient, and bias.

        Furthermore, we performed a detailed scalability analysis on \texttt{Qwen-3} models of different sizes--\texttt{1.7}, \texttt{4}, \texttt{8}, and \texttt{32B}--to assess the consistency of our method across model scales. Our experiments were conducted on NVIDIA A40 (46 GB memory) and H200 (140 GB memory) GPUs for larger LLMs. For simplicity of implementation, AAI was applied only to reweight attention scores during the input-prompting (prefill) phase. We employed greedy decoding, selecting the highest-logit token at each generation step to produce deterministic outputs.
         
    \subsection{Experimental Results}
        
\begin{figure*}[ht]
        \centering
        \footnotesize
        \centerline{
        \begin{tikzpicture}
            \begin{axis}[width=0.33\linewidth,height=0.23\linewidth,
                axis lines=middle,
                ymin=65, ymax=86,
                xmin=-2, xmax=34, 	          
                x label style={at={(current axis.right of origin)},anchor=south, above=1mm},
                legend style ={at={(0.7,0.8)}, anchor=north, align=left},
                legend cell align={left},
                legend style={ anchor=north,legend columns=-1},
                legend to name={legend_ratio},
                xlabel= \#params ($B$),
                ylabel=\textit{ProofWriter Acc.},
                ylabel style = {at={(0.06,1.1)},},
                xticklabel,  
                enlargelimits = false,
                xticklabels from table={./pw_ab.tex}{depth},
                every axis plot/.append style={ thick},
                xtick=data]

                \addplot [ opacity=0.75,mark=*, color=orange,  error bars/.cd, y dir=both, y explicit,] table [ x=idx, y=CoT, col sep=space, 
                    ]   {./pw_ab.tex};
                \addlegendentry{CoT} 
                \addplot [ opacity=0.75,mark=triangle*, 
                color=blue,  error bars/.cd, y dir=both, y explicit,] table [ x=idx, y=LogicCoT1, col sep=space, 
                    ]   {./pw_ab.tex};
                \addlegendentry{SymbA CoT}  
                \addplot [ opacity=0.75,mark=star, color=Cyan,  error bars/.cd, y dir=both, y explicit,] table [ x=idx, y=AAVertical, col sep=space, 
                    ]   {./pw_ab.tex};
                \addlegendentry{AAI$^{agg.}$} 
                \addplot [ opacity=0.9,mark=diamond*, color=red,  error bars/.cd, y dir=both, y explicit,] table [ x=idx, y=AADiagonal, col sep=space, 
                    ]   {./pw_ab.tex};
                \addlegendentry{AAI} 
                
            \end{axis} 
        \end{tikzpicture}
        
        \begin{tikzpicture}
            \begin{axis}[width=0.33\linewidth,height=0.23\linewidth,
                axis lines=middle,
                ymin=93, ymax=101,
                xmin=-2, xmax=34, 	          
                x label style={at={(current axis.right of origin)},anchor=south, above=1mm},
                legend style ={at={(0.7,0.8)}, anchor=north, align=left},
                legend cell align={left},
                legend style={ anchor=north,legend columns=-1},
                legend to name={legend_ratio_3},
                xlabel= \#params ($B$),
                ylabel=\textit{ProntoQA Acc.},
                ylabel style = {at={(0.06,1.1)},},
                xticklabel,  
                enlargelimits = false,
                xticklabels from table={./prontoQA_ab.tex}{depth},
                every axis plot/.append style={ thick},
                xtick=data]

                \addplot [ opacity=0.75,mark=*, color=orange,  error bars/.cd, y dir=both, y explicit,] table [ x=idx, y=CoT, col sep=space, 
                    ]   {./prontoQA_ab.tex};
                \addlegendentry{CoT} 
                \addplot [ opacity=0.75,mark=triangle*, 
              color=blue,  error bars/.cd, y dir=both, y explicit,] table [ x=idx, y=LogicCoT, col sep=space, 
                    ]   {./prontoQA_ab.tex};
                \addlegendentry{SymbA CoT} 
                \addplot [ opacity=0.9,mark=star, color=Cyan,  error bars/.cd, y dir=both, y explicit,] table [ x=idx, y=AAVertical, col sep=space, 
                    ]   {./prontoQA_ab.tex};
                \addlegendentry{AAI$^{agg.}$} 
                \addplot [ opacity=0.9,mark=diamond*, color=red,  error bars/.cd, y dir=both, y explicit,] table [ x=idx, y=AADiagonal, col sep=space, 
                    ]   {./prontoQA_ab.tex};
                \addlegendentry{AAI} 
            \end{axis} 
        \end{tikzpicture}
        \begin{tikzpicture}
            \begin{axis}[width=0.33\linewidth,height=0.23\linewidth,
                axis lines=middle,
                ymin=72, ymax=93,
                xmin=-2, xmax=34, 	          
                x label style={at={(current axis.right of origin)},anchor=south, above=1mm},
                legend style ={at={(0.7,0.8)}, anchor=north, align=left},
                legend cell align={left},
                legend style={ anchor=north,legend columns=4},
                legend to name={legend_ratio_2},
                xlabel= \#params ($B$),
                ylabel=\textit{LogicDeduction Acc.},
                ylabel style = {at={(0.06,1.1)},},
                xticklabel,  
                enlargelimits = false,
                xticklabels from table={./ld_ab.tex}{depth},
                every axis plot/.append style={ thick},
                xtick=data]

                \addplot [ opacity=0.75,mark=*, color=orange,  error bars/.cd, y dir=both, y explicit,] table [ x=idx, y=CoT, col sep=space, 
                    ]   {./ld_ab.tex};
                \addlegendentry{CoT} 
                \addplot [ opacity=0.75,mark=triangle*, 
              color=blue,  error bars/.cd, y dir=both, y explicit,] table [ x=idx, y=LogicCoT, col sep=space, 
                    ]   {./ld_ab.tex};
                \addlegendentry{SymbA CoT} 
                \addplot [ opacity=0.75,mark=star, color=Cyan,  error bars/.cd, y dir=both, y explicit,] table [ x=idx, y=AAVertical, col sep=space, 
                    ]   {./ld_ab.tex};
                \addlegendentry{AAI$^{agg.}$} 
                \addplot [ opacity=0.9,mark=diamond*, color=red,  error bars/.cd, y dir=both, y explicit,] table [ x=idx, y=AADiagonal, col sep=space, 
                    ]   {./ld_ab.tex};
                \addlegendentry{AAI}  
            \end{axis} 
        \end{tikzpicture}
        }
        
        \ref{legend_ratio}
        \caption{Performance across different model sizes of \texttt{Qwen-3} with three prompting techniques on three datasets. 
        \label{fig_modelsize}}
    \end{figure*}
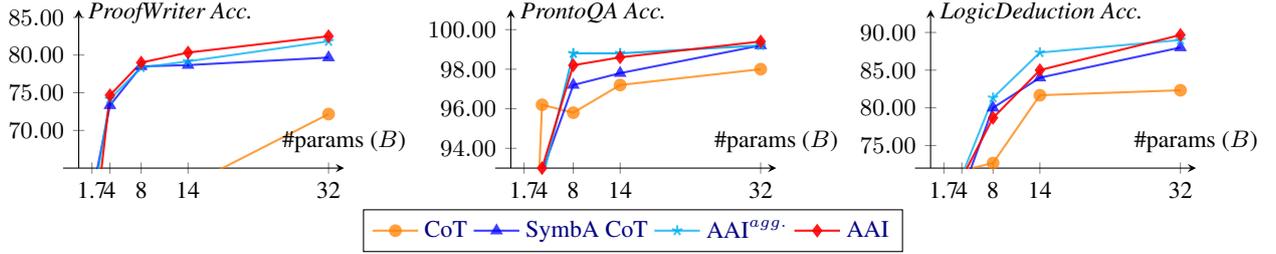

        \paragraph{Main results.} We evaluate the performance of the proposed method from two perspectives: (1) scalability with respect to model size (Fig.~\ref{fig_modelsize}), and (2) generalization across different LLMs (Table~\ref{tab_mainresult}). 
        In Table~\ref{tab_mainresult}, the proposed method demonstrates the generalization capability. Specifically, it yields improvements of $+2.5$ and $+3.0$ accuracy on the ProofWriter and LogicalDeduction datasets, respectively, when applied to the \texttt{OLMo-2 32B} model. Moreover, when using \texttt{Phi-4}, AAI achieves an additional $+0.66$ accuracy gain on the LogicalDeduction dataset. 
        
        As shown in Fig.~\ref{fig_modelsize}, by using \texttt{Qwen-3} models, our method (AAI -- represented by the red diamond line) exhibits increasingly stronger gains as the model size grows--particularly on the ProofWriter and Logical Deduction datasets--achieving notable improvements of $+2.83$ and $+1.67$ accuracy, respectively, over the baseline (SymbolA.CoT). For the ProntoQA dataset, the performance reaches near saturation with the $32$B model (over $99\%$ accuracy); hence, the improvement gains are more pronounced for the 8B and 14B model sizes.
        
        Our AAI mechanism also demonstrates strong generalization across diverse structured prompting strategies and domains. It consistently improves performance across datasets and LLMs when combined with the \textit{Compact Symbolic-Aided CoT} prompting and remains effective on GSM8k, a mathematical reasoning benchmark (Table~\ref{tab_mainresult}). Notably, under the GSM8k prompting setting (Table~\ref{tab_compact_prompting_GSM8k} in Appendix~\ref{apd_prompting}), few-shot prompting enables LLMs to autonomously decompose the input context into rule-like structures. 
        The key principle underlying the application of our AAI mechanism is to decompose the overall problem into subproblems and to appropriately steer the attention of LLMs to address each subproblem through a structured reasoning process. These results highlight the potential applicability of our approach to other reasoning tasks, such as commonsense and multi-hop reasoning.  
        
        \begin{figure}[htbp]
                \centering
                \begin{tikzpicture}
                \tikzstyle{every node}=[font=\small]
                    \begin{axis} [ybar,
                        width=1.\linewidth,height=0.45\linewidth,
                        bar width = 9pt,
                        ymin=64,
                        tick label style={font=\small},
                        ylabel=ProofWriter Acc. ,
                         x label style={at={(current axis.right of origin)},anchor=south, right=4mm, below=1mm}, 
                         y label style={at={(current axis.left of origin)},anchor=south, left=-8mm, below=-14mm}, 
                        ybar=-22, %
                        enlarge x limits = 0.25,
                        symbolic x coords = {Qwen3 32B,OLMo 32B,Phi-4 14B},
                        legend style={
                            at={(0.4,1.5)},
                           legend columns=-1,
                           anchor=north,
                           /tikz/every node/.style={anchor=west}
                       },
                        ybar,
                        x tick label style={
                            /pgf/number format/1000 sep=},
                         xticklabel style={rotate=0,anchor=center, yshift=-5pt},
                          minor y tick num=4,
                        ymajorgrids,
                        ]
                        \addlegendentry{SymbA.CoT} 
                        \addplot[ybar, bar shift=-17pt, fill=blue1,   ] coordinates{(Qwen3 32B, 79.67) (OLMo 32B, 65.00) (Phi-4 14B, 76.33)};
                        
                        \addlegendentry{AAI$^{\mathrm{-HeadsSel.}}$} 
                        \addplot[ybar, bar shift=-6pt,  fill=blue2,  postaction={pattern=dots}] coordinates{(Qwen3 32B, 82.67) (OLMo 32B,68.50) (Phi-4 14B,74.83)};
                        
                        \addlegendentry{AAI$^{\mathrm{agg.}}$} 
                        \addplot[ybar, bar shift=5pt,   fill=blue3,  postaction={pattern=grid}] coordinates{(Qwen3 32B,81.83) (OLMo 32B, 67.33) (Phi-4 14B,75.17)};
                        
                        \addlegendentry{AAI} 
                        \addplot[ybar, bar shift=16pt,  fill=blue4,    postaction={pattern=north west lines}] coordinates{(Qwen3 32B, 82.50) (OLMo 32B,67.50) (Phi-4 14B, 76.33)}; 
                    \end{axis} 
                    \end{tikzpicture}
                \begin{tikzpicture}
                \tikzstyle{every node}=[font=\small]
                    \begin{axis}[
                        ybar,
                        ymin=45, ymax=98,
                        axis y discontinuity=parallel,   
                        enlarge y limits=false,
                        width=1.\linewidth, height=0.45\linewidth,
                        bar width=9pt,
                        ytick={60,88.},
                        tick label style={font=\small},
                        ylabel=L.Deduction Acc.,
                        x label style={at={(current axis.right of origin)},anchor=south, right=4mm, below=1mm},
                        y label style={at={(current axis.left of origin)},anchor=south, left=-8mm, below=-14mm},
                        enlarge x limits=0.25,
                        symbolic x coords={Qwen3 32B, OLMo 32B, Phi-4 14B},
                        legend style={
                            at={(0.5,1.15)},
                            anchor=south,
                            legend columns=2,
                            /tikz/every node/.style={anchor=west}
                        },
                        xticklabel style={rotate=0,anchor=center,yshift=-5pt},
                        ymajorgrids,
                        yminorgrids,
                    ]
                
                    \addplot[ybar, bar shift=-17pt, fill=blue1] 
                        coordinates{(Qwen3 32B, 88) (OLMo 32B, 51.33) (Phi-4 14B, 91.67)};
                    \addplot[ybar, bar shift=-6pt, fill=blue2, postaction={pattern=dots}] 
                        coordinates{(Qwen3 32B, 90.67) (OLMo 32B, 51.33) (Phi-4 14B, 91.33)};
                    \addplot[ybar, bar shift=5pt, fill=blue3, postaction={pattern=grid}] 
                        coordinates{(Qwen3 32B, 89.00) (OLMo 32B, 54.00) (Phi-4 14B, 91.67)};
                    \addplot[ybar, bar shift=16pt, fill=blue4, postaction={pattern=north west lines}] 
                        coordinates{(Qwen3 32B, 89.67) (OLMo 32B, 54.33) (Phi-4 14B, 92.33)};
                \end{axis}
                \end{tikzpicture}
            \caption{Ablation study on our AAI method. }
            \label{fig_ablation}
            \end{figure}
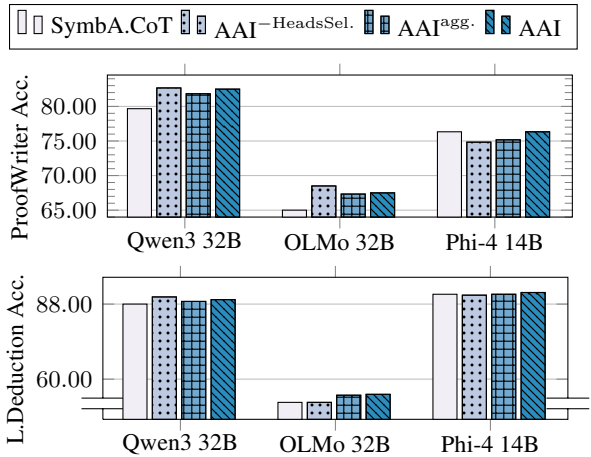

    \begin{table*}[!t]
        \small
            \caption{Performance comparison between our method and existing approaches. \textit{E.Solver} denotes systems that incorporate an external symbolic solver module. Results marked with $^*$ are reproduced or obtained in this work. The \textbf{best} performance in each category is highlighted.\label{tab_mainresult}\vspace{-5pt}}
        \centering
        \resizebox{\textwidth}{!}{%
            \begin{tabular}{llrrrrrrr}
                \toprule  \multirow{1}{*}{\textbf{Methods + LLMs}}&\textbf{Interaction mode }&\textbf{ProofWriter}&
                \textbf{ProntoQA} &\textbf{L.Deduction}&\textbf{FOLIO} &\textbf{GSM8k}     \\
                
                \midrule
                 \multicolumn{4}{l}{\quad \quad\underline{\textit{Closed-source LLM: GPT-4}}} \\
                 
                CoT~\cite{sun-etal-2024-determlr}  & Non-Interactive & $67.41$ &   $91.00$ & $73.33$ &$67.65$ &--\\
                Logic-LM~\cite{pan-etal-2023-logic}  & Interactive +E.Solver & {$79.66$} & $83.20$& {$87.63$} &$78.92$ &--\\
                DetermLR~\cite{sun-etal-2024-determlr}  & Interactive +Program& $79.17$ &  $98.60$ & $85.00$&$75.49$&--\\
                SymbCoT~\cite{xu-etal-2024-faithful}  & Interactive +Program& $\mathbf{82.50}$ &   {$99.60$}&$\mathbf{93.00}$ &$\mathbf{83.33}$&--\\
                {Symbolic-Aided CoT \citep{nguyen2025noniterative}} & Non-Interactive & $77.09$  &   $\boldsymbol{100.00}$ & $86.33$&$74.51$&--\\ 

                \midrule 

                \multicolumn{4}{l}{\quad \quad\underline{\textit{Qwen3-32B}}} \\
                {Symbolic-Aided CoT} \cite{nguyen2025noniterative}$^*$ & Non-Interactive & $\valOfPW{4}{LogicCoT}$ & ${\valOfProntoQA{4}{LogicCoT}}$& ${\valOfLD{4}{LogicCoT}}$&$69.61$&--\\    
                \rowcolor{gray!20}
                { +AAI (ours)}$^*$ & Non-Interactive & $\mathbf{\valOfPW{4}{AADiagonal}}$ & $\mathbf{\valOfProntoQA{4}{AADiagonal}}$& $\mathbf{\valOfLD{4}{AADiagonal}}$&${71.09}$&--\\   
                {Compact Symbolic-Aided CoT (ours)}$^*$  & Non-Interactive & $74.00$&	$91.20$&	$74.67$&	$70.59$&$91.81$\\    
                \rowcolor{gray!20}
                { +AAI (ours)}$^*$ & Non-Interactive & ${74.83}	$&	${95.20}$&	$	{75.33}$&	$	\mathbf{75.00}$&$\mathbf{92.27}$\\   
                 \midrule                
                \multicolumn{4}{l}{\quad \quad\underline{\textit{OLMo-2-0325-32B-Instruct}}} \\
                {Symbolic-Aided CoT} \cite{nguyen2025noniterative}$^*$& Non-Interactive & $\valOfPWOlmo{4}{LogicCoT}$ & ${\valOfProntoQAOlmo{4}{LogicCoT}}$& ${\valOfLDOlmo{4}{LogicCoT}}$&$\mathbf{64.22}$&--\\    
                
                \rowcolor{gray!20}
                { +AAI (ours)}$^*$ & Non-Interactive & $\mathbf{\valOfPWOlmo{4}{AADiagonal}}$ &  $\mathbf{\valOfProntoQAOlmo{4}{AADiagonal}}$ &  $\mathbf{\valOfLDOlmo{4}{AADiagonal}}$&$\mathbf{64.22}$&--\\   
                {Compact Symbolic-Aided CoT (ours)}$^*$  & Non-Interactive & $55.17$&	$	{73.60}$&	$	45.00$&	$	52.94$&$82.94$\\    
                \rowcolor{gray!20}
                { +AAI (ours)}$^*$ & Non-Interactive & ${56.00}$&	$	72.80$&	$	{46.00}$&	$	{53.92}$&$\mathbf{83.02}$\\   
                 \midrule                
                \multicolumn{4}{l}{\quad \quad\underline{\textit{Phi-4}} \textit{(14B)}} \\
                {Symbolic-Aided CoT} \cite{nguyen2025noniterative}$^*$& Non-Interactive & $\mathbf{\valOfPWphiF{4}{LogicCoT}}$ & $\mathbf{\valOfProntoQAphiF{4}{LogicCoT}}$& ${\valOfLDphiF{4}{LogicCoT}}$&$70.59$&--\\   
                
                \rowcolor{gray!20}
                { +AAI (ours)}$^*$ & Non-Interactive & $\mathbf{\valOfPWphiF{4}{AADiagonal}}$ &  ${\valOfProntoQAphiF{4}{AADiagonal}}$& $\mathbf{\valOfLDphiF{4}{AADiagonal}}$&$\mathbf{72.06}$&--\\   
                {Compact Symbolic-Aided CoT (ours)}$^*$  & Non-Interactive & $66.00$&	$	97.00$&	$	79.33$&	$	64.21$&$87.41$\\    
                \rowcolor{gray!20}
                { +AAI (ours)}$^*$ & Non-Interactive & $67.00$&	$	97.80$&	$	80.00$&	$	64.71$&$\mathbf{87.57}$\\   

                \bottomrule 
                 
            \end{tabular} 
            } 
    \end{table*}

    \paragraph{Ablation studies.}     
        To evaluate the contribution of each sub-component in our proposed method, including the effects of Attention Head selection and reweighting, we conducted ablation studies on two datasets that are not affected by saturation performance: ProofWriter and LogicalDeduction (Fig.~\ref{fig_ablation}).
    In comparison, the AAI method tends to select copy heads and anchor heads, while AAI$^{\mathrm{agg.}}$ tends to select aggregation heads. Meanwhile, AAI$^{\mathrm{-Heads\,Sel.}}$ applies attention reweighting to all heads without selection. 
    Notably, in the case of the \texttt{Qwen-3} and \texttt{OLMo} $32$B model, our methods attained consistently higher performance across all three configurations, highlighting the effectiveness of our attention reweighting mechanism in embedding prior knowledge within LLMs.
    
    We found that using copy heads and anchor heads leads to more consistent performance improvements over the baseline model across various LLMs and datasets.
    In the AAI$^{\mathrm{agg.}}$ variant, the proportion of selected heads typically ranges from $30$-$40\%$ of all attention heads, whereas the standard AAI selects $7$-$20\%$. Because both AAI$^{\mathrm{-Heads\,Sel.}}$ and AAI$^{\mathrm{agg.}}$ affect a substantially larger portion of aggregation-related attention heads, they are more likely to alter the LLM’s decoding states. This broader intervention can yield noticeable performance gains (\textit{e.g.,} the 8–14B region in Fig.~\ref{fig_modelsize}), but it also entails a higher risk of degrading the model’s overall behavior. In contrast, our main AAI mechanism targets anchor or copy heads, which are responsible for storing information, as the attention reweighting strategy is designed to relocate referencing information.
    This targeted intervention offers a more controlled and interpretable adjustment, whereas reweighting a large fraction of aggregation attention heads may introduce instability.
        
\begin{figure*}[ht]
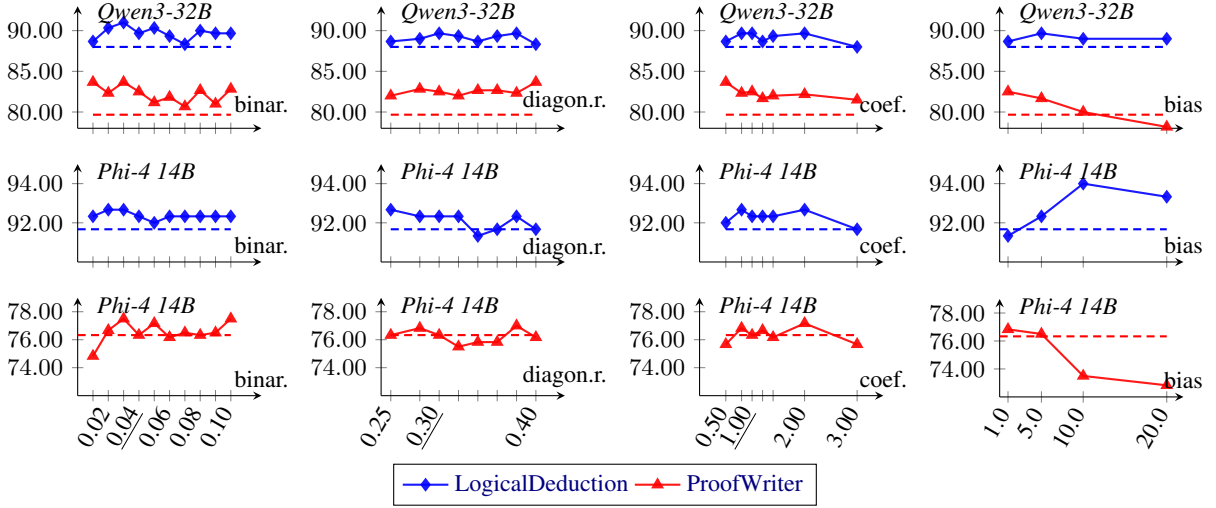

            \centering
            \footnotesize
            \centerline{
            

            
            }
            \ref{legend_binrate}
            \caption{Performance (accuracy) changes with respect to different hyperparameter values  (attention binarization threshold, $s^{diagonal}$, reweighting coefficient ($c$), and bias constant ($b$) in Eq.~\ref{eq_reweighting}) on two benchmark datasets, Logical Deduction and ProofWriter.  The dashed line denotes the baseline system, which does not use our AAI mechanism. Values marked with underscores indicate the default settings used in Table~\ref{tab_mainresult} ($binar.=0.04$, $diagon.r.=0.3$, $ coef.=1.0$, $bias=0$). In the last column, the coefficient is set to $coef.=0$ and the bias value is varied.  \label{fig_hyper_thres_change}}
        \end{figure*}

    \paragraph{Hyperparameter sensitivity analysis.}
    To assess the robustness of our AAI method with respect to configuration hyperparameters, we conducted extensive experiments on the LogicalDeduction and ProofWriter datasets using two LLMs, \texttt{Qwen3-32B} and \texttt{Phi-4} (Fig.~\ref{fig_hyper_thres_change}). The evaluated hyperparameters include the binarization threshold, diagonal criterion, reweighting coefficient, and bias values.
    
    The results demonstrate that our AAI method is robust to variations in configuration hyperparameters. In addition, both \texttt{Qwen3-32B} and \texttt{Phi-4} show consistent performance improvements on the LogicalDeduction dataset compared to the baseline. In the rightmost column of the figure, by setting the \textit{coef} value to 0, a constant term (\textit{bias}), which is not adaptive to the original attention values, is added to the attention matrix. As the magnitude of this constant increases, the performance of the LLMs can be adversely affected, particularly on datasets with strong rule dependencies such as ProofWriter (the red lines of the right most column).
    For experiments with the \texttt{Phi-4 14B} model on the ProofWriter dataset (the bottom row), only a small difference is observed compared with the baseline model.  
    Our observations indicate that a large portion of the errors stem from incorrect rule selection or mismatches between the question and the premises in the KB set during the final answer step \citep{nguyen2025noniterative}. Consequently, overall performance improvements remain limited, despite our AAI method supporting more faithful reasoning through improved rule representations. We leave addressing these issues to future work.
        

            
\begin{figure*}[!htbp]
        \centering 
        \includegraphics[width=0.32\linewidth, keepaspectratio, 
                trim={1.2cm 0.5cm 1.5cm 1.1cm  }, page=1, clip=true]{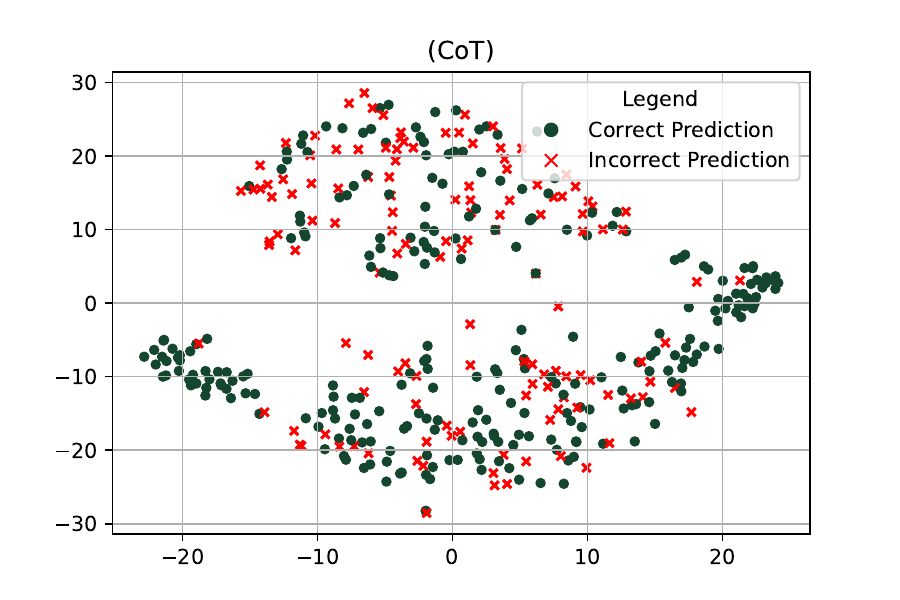}
        \includegraphics[width=0.32\linewidth, keepaspectratio, 
                trim={1.2cm 0.5cm 1.5cm 1.1cm }, page=1, clip=true]{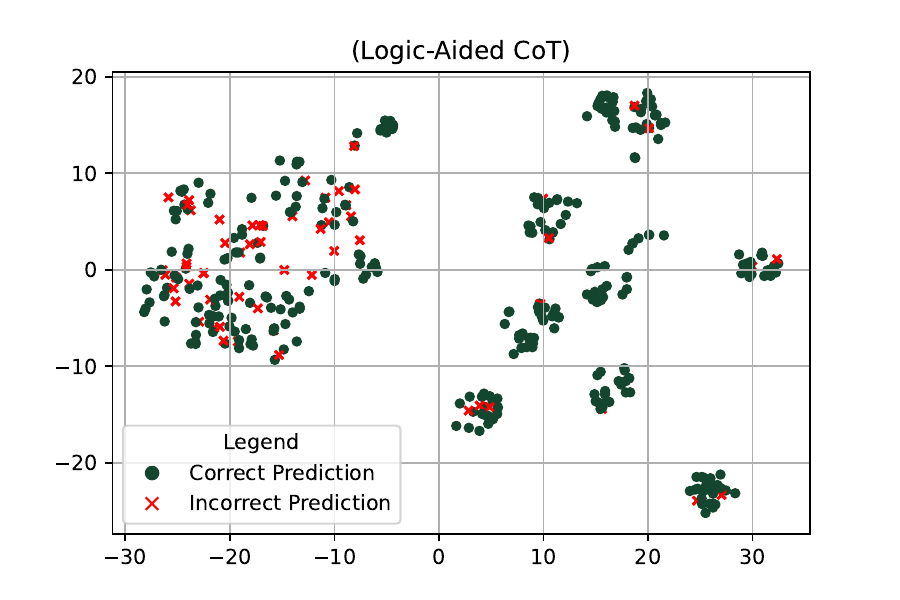}
        \includegraphics[width=0.32\linewidth, keepaspectratio, 
                trim={1.2cm 0.5cm 1.5cm 1.1cm }, page=1, clip=true]{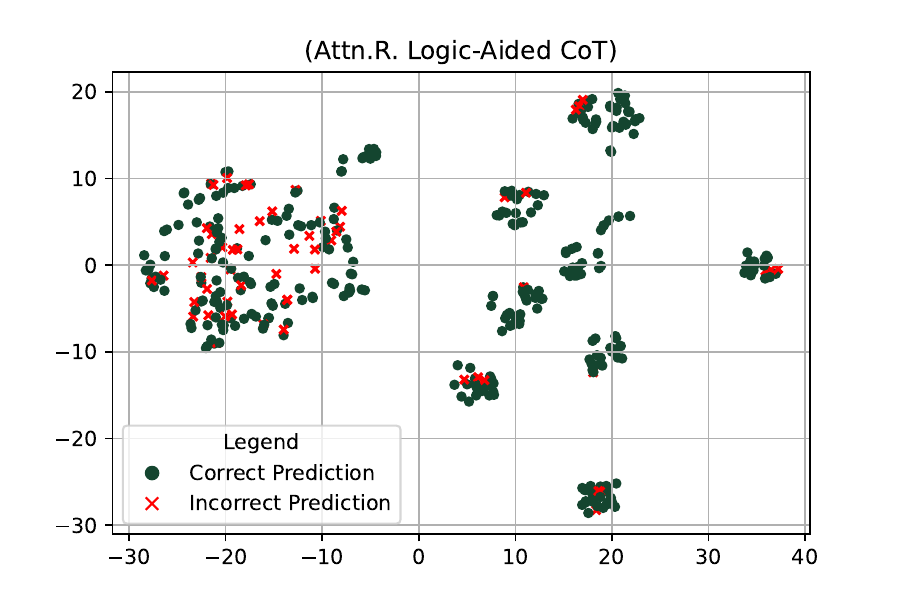}
       
        \caption{Visualization of the last hidden states of a sample encoded by \texttt{Qwen3-8B} LLM from the ProofWriter dataset under three reasoning settings: standard CoT, Symbolic-Aided CoT, and our proposed Attention-Aware Intervention (AAI), shown from left to right, respectively.
        }\label{fig_viz_last_word_vects}
    \end{figure*}

    \paragraph{Semantic representation.} Fig.~\ref{fig_viz_last_word_vects} presents a t-SNE visualization of the last-layer hidden states of the final token in the input prompt, obtained using the \texttt{Qwen3-8B} model on the ProofWriter dataset, comparing standard CoT, Symbolic-Aided CoT, and our proposed AAI method. We observe that both Symbolic-Aided CoT and AAI exhibit similar distribution patterns, segmenting the dataset into finer-grained sub-clusters compared to standard CoT. Moreover, misclassified samples (red points) tend to appear in regions where clusters overlap. These results indicate that both Symbolic-Aided CoT and AAI effectively incorporate structural information into their representations. When comparing Symbolic-Aided CoT with our AAI method, AAI produces more distinct and compact clusters, as reflected by the wider range of values along both the x- and y-axes and the denser grouping of points within each cluster.
    
    \paragraph{Attention head pattern.} In 
    Fig.~\ref{fig_attn_pattern_full} Appendix~\ref{apd_attention_head_pattern}, we present the attention head pattern scores for all heads across $36$ layers in the \texttt{Qwen3-8B} LLM. Each attention head is evaluated using three metrics: diagonal, vertical, and horizontal scores. The red zoomed-in region on the left highlights a head with a high diagonal score. The top-right blue zoomed-in region illustrates heads exhibiting high values across all three scores. In contrast, the bottom blue zoomed-in region shows heads characterized by a high vertical score but relatively low horizontal and diagonal scores. The results show that a large proportion of heads in this model exhibit a high rate of vertical patterns. In contrast, the lower layers typically show high scores across all three pattern types--vertical, horizontal, and diagonal--indicating that information in these layers is beginning to be integrated and features are being extracted based on more complex dependencies.

   

\section{ Related Work\label{sec_related_work}} 

\paragraph{Logical reasoning.}  Existing research on logical reasoning with LLMs generally follows two directions. The first leverages external symbolic solvers, where LLMs translate natural language into formal logic and delegate inference to symbolic reasoners~\citep{pan-etal-2023-logic,ye2023satlm,olausson-etal-2023-linc,pmlr-v202-gao23f,yao2023react,xu-etal-2024-symbol}. In contrast, our approach embeds structure directly into the reasoning process, enabling the LLM itself to act as a symbolic reasoner without relying on external solvers. The second direction focuses on interactive prompting frameworks, such as least-to-most prompting, which decompose complex reasoning into subtasks (\textit{e.g.,} rule selection, premise derivation, and scoring) to improve accuracy and faithfulness~\citep{zhou2023leasttomost,zhang2025cumulative,kazemi-etal-2023-lambada,feng-etal-2024-language,sun-etal-2024-determlr,lee-hwang-2025-symba}. 
Unlike these methods, our approach is non-interactive, producing the final answer in a single forward pass through the LLM. Compared to \citet{nguyen2025noniterative}, we further enhance the explainability of Symbolic-aided CoT prompting and introduce an Attention-Aware Intervention mechanism to inject prior knowledge into LLMs.

\paragraph{Attention analysis.}
Recent work has examined the interpretability of Transformer attention heads by analyzing their functional roles and activation patterns. Prior studies have identified causally relevant or reasoning-specialized heads through interventions~\citep{nam2025causalheadgatingframework,park2025thinkingsparksemergentattention}. \citet{elhelo-geva-2025-inferring} and \citet{zhao-etal-2025-analyzing} examined how attention parameters or activation stability relate to task generalization. \citet{bogdan2025thoughtanchorsllmreasoning} explore sentences as thought anchors and analyze patterns of attention heads with high relevance. Other approaches steer attention to emphasize user-specified instruction components~\citep{zhang2024tell,spotlight_instruct}.
In contrast, our work investigates attention behavior under structured reasoning prompts with explicit symbolic operators. We introduce a fine-grained taxonomy of anchor, aggregation, and copy heads, and propose a dynamic attention reweighting mechanism to inject prior knowledge, bridging attention-level interpretability with controllable symbolic reasoning in non-interactive settings.

\section{Conclusion\label{sec_conclusion}} 
We presented Attention-Aware Intervention, a non-interactive and end-to-end framework for enhancing logical reasoning with LLMs. By incorporating symbolic structural information into the few-shot prompt, we revealed that special attention heads exhibit activation patterns aligned with logical reasoning operators. Leveraging this insight, AAI modulates attention scores across these heads to guide reasoning without relying on external components or complex multi-step interactions. 
Our findings shed light on the development of scalable, interpretable reasoning systems with LLMs.

\section*{Limitations}
While our proposed Attention-Aware Intervention (AAI) framework demonstrates strong performance and improved interpretability across multiple logical reasoning benchmarks, it also has several limitations. First, our method has been primarily evaluated on synthetic and small-to-medium-scale datasets, which may not fully capture the complexity of real-world reasoning scenarios. Second, although AAI effectively injects symbolic structure into LLMs, the intervention depends on model-specific attention patterns, which may limit its generalizability to architectures that do not employ the self-attention mechanism. Finally, our current work focuses exclusively on logical reasoning tasks; extending AAI to other domains, such as commonsense reasoning, remains an important direction for future research.

\section*{Acknowledgments}

This work is partially supported by the Nakajima Foundation and JSPS KAKENHI Grant Number 22H00524. We used ABCI 3.0 provided by AIST and AIST Solutions with support from ``ABCI 3.0 Development Acceleration Use''.


\bibliography{custom}

\appendix


\section{AI Usage Declaration }

AI tools were used for grammar checking and formatting of tables and figures, and polish writing. All technical content
and implementations were written by the authors.

\section{Attention Head Observation\label{apd_attn_map_explain}}
    
\begin{figure*}[!htbp]
        \centering 
        \includegraphics[height=1.45\linewidth, keepaspectratio, 
                trim={0 0 0 0}, page=1, clip=true]{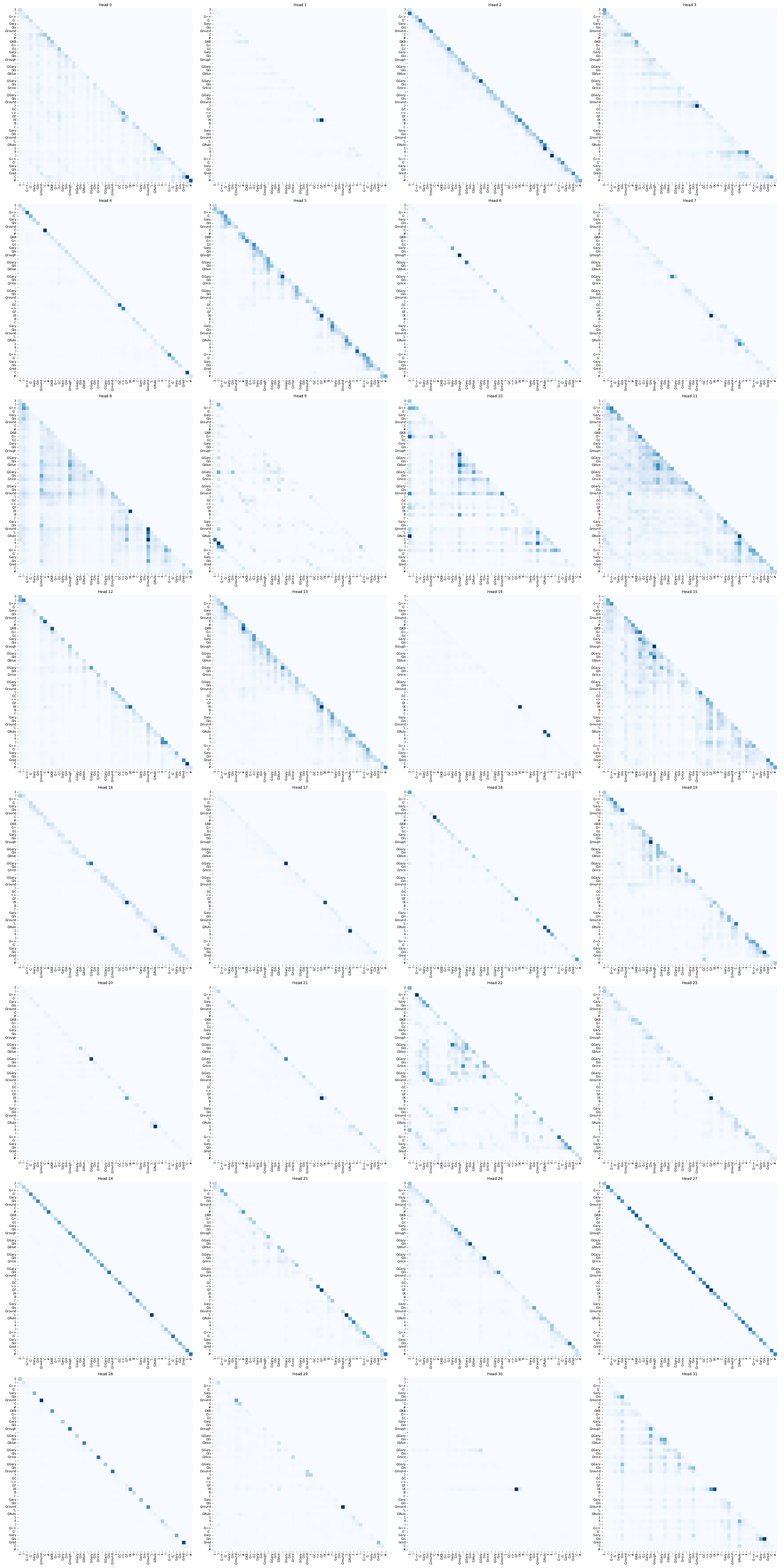}
        \caption{Visualization of attention heads in the \texttt{Qwen3-8B} model on an example from the ProofWriter dataset.}\label{fig_attention_prelim_exp_full}
    \end{figure*}

    In the self-attention layer of the Transformer architecture, for a token at position $t$ in a sentence of length $L$, we have the corresponding query ($q_t \in \mathbb{R}^{1\times d_k}$), key ($k_t \in \mathbb{R}^{1\times d_k}$) and value ($v_t \in \mathbb{R}^{1\times d_v}$) representations. Intuitively, the representation of the token at step $t$ ($h_t$) is computed from the value vectors of previous tokens ($v_{\leq t} \in \mathbb{R}^{t\times d_v}$), weighted by their similarity to the current token as determined by the $\textrm{softmax}$ function: 
 \begin{equation}
     h_t = \textrm{softmax}\left( \frac{q_t   k_{\leq t}^\top  }{\sqrt{d_k}}\right)  v_{\leq t}
 \end{equation}
Fig.~\ref{fig_attention_prelim_exp_full} (full-scale version) visualizes the attention heads of the \texttt{Qwen3-8B} model on a sample from the ProofWriter dataset. Higher attention scores are indicated by darker colors.  
Based on this mechanism, we identify three characteristic attention patterns:
\textbf{(Anchor Head)}: When a token at position $t$ attends almost exclusively to itself (close to $100$\%), only the main diagonal elements of the attention map are activated. This pattern is observed in token–attribute representations in Head28 (Fig.\ref{fig_attention_prelim_exp}), for example, \textit{``blue''} and \textit{``rough''}.
\textbf{(Aggregation Head)}: When a sequence of tokens attends strongly to a specific earlier token in order to incorporate information introduced by that token, an aggregation pattern emerges. For example, after the token \textit{``round''} is generated, this information propagates to many subsequent tokens (Head8, Fig.\ref{fig_attention_prelim_exp}). In the attention map, this behavior appears as a \textit{vertical line}. Conversely, when a query token needs to gather information from multiple previous tokens, the pattern appears as \textit{horizontal lines}.
\textbf{(Copy Head)}:  A diagonal pattern emerges when a span of generated tokens is copied forward. Specifically, if a span of three tokens starting at position $k$ is copied to position $h$ ($h > k$), the attention scores are high at the token pairs ($k \rightarrow h$), ($k+1 \rightarrow h+1$), ($k+2 \rightarrow h+2$). This pattern is observed in Head22 (Fig.\ref{fig_attention_prelim_exp}), where the span \textit{``Gary is round''} is copied from its original generation location to the \textit{KB variable set} region in Symbolic-Aided CoT prompting.

\section{Improving examples}
\begin{figure*}[!htbp]
    \centering 
    \centerline{
    \includegraphics[width=\linewidth, keepaspectratio, 
            trim={0cm  0 0 0 }, page=1, clip=true]{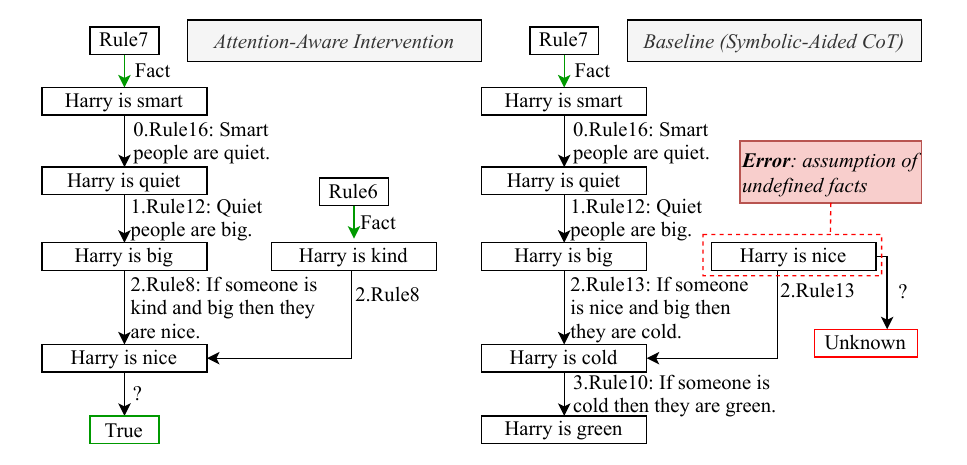} 
            }
   
    \caption{An example illustrating the improvement achieved by our AAI (left) compared to the baseline method (right) on the ProofWriter dataset.
    }\label{fig_improving_sample_pw1}
\end{figure*}

\begin{figure*}[!htbp]
    \centering 
    \centerline{
    \includegraphics[width=\linewidth, keepaspectratio, 
            trim={0cm  0 0 0 }, page=1, clip=true]{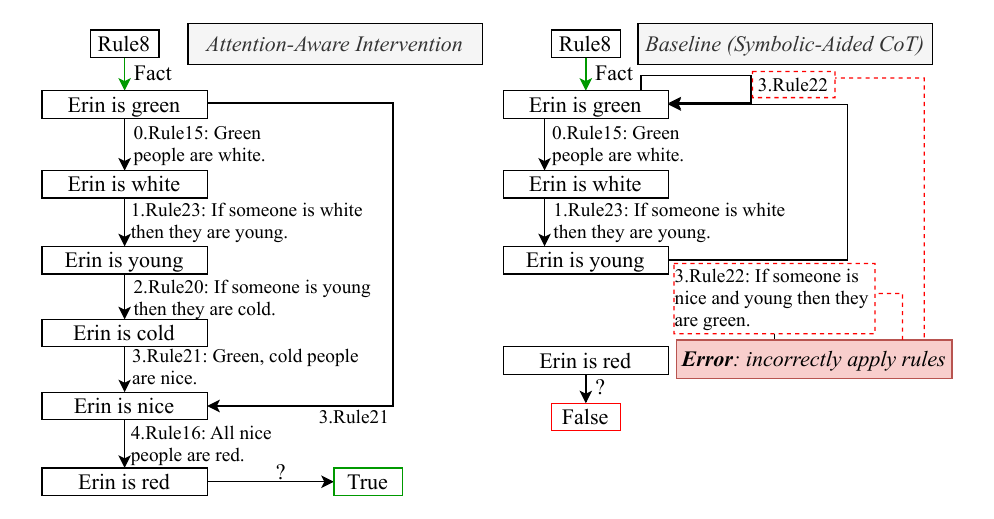} 
            }
   
    \caption{An example illustrating the improvement achieved by our AAI (left) compared to the baseline method (right) on the ProofWriter dataset.
    }\label{fig_improving_sample_pw2}
\end{figure*}

\begin{figure*}[!htbp]
    \centering 
    \centerline{
    \includegraphics[width=\linewidth, keepaspectratio, 
            trim={0cm  0 0 0 }, page=1, clip=true]{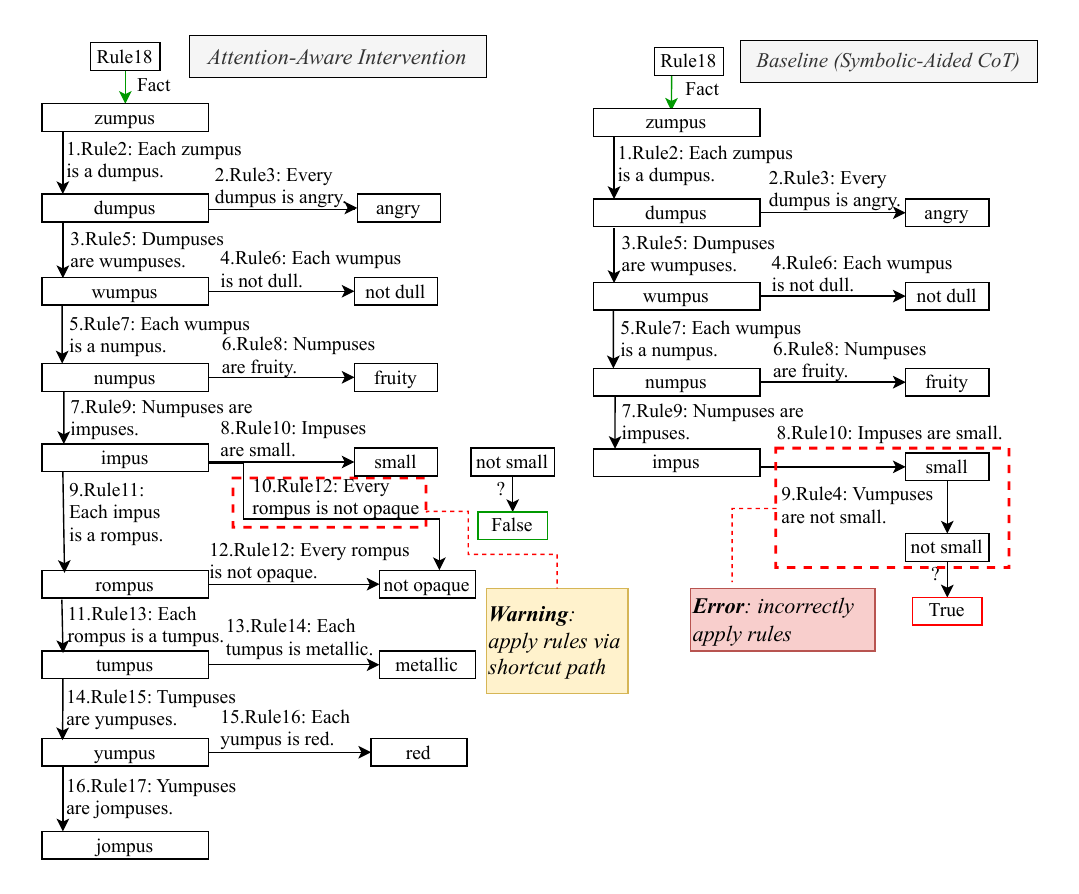} 
            }

    \caption{An example illustrating the improvement achieved by our AAI (left) compared to the baseline method (right) on the ProntoQA dataset. 
    }\label{fig_improving_sample_pronto}
\end{figure*}

In this section, we present several representative cases in which our AAI mechanism improves the baseline system on the ProofWriter dataset (Figures~\ref{fig_improving_sample_pw1} and~\ref{fig_improving_sample_pw2}) and the ProntoQA dataset (Figure~\ref{fig_improving_sample_pronto}).
The raw prompting outputs shown in Table~\ref{tab_prompting_PW_impr_sample} are aligned with the corresponding proof tree in Figure~\ref{fig_improving_sample_pw1}.
The improvements achieved by our AAI mechanism typically arise from correcting incorrect rule selection and improper rule application during the reasoning process.

\begin{table}[htbp]
\small
\centering
\caption{Our Symbolic-Aided CoT + AAI prompting output for an example from the ProofWriter dataset (ID: \texttt{ProofWriter\_AttNoneg-OWA-D5-1189\_Q7}).
\label{tab_prompting_PW_impr_sample} 
}
\resizebox{\linewidth}{!}{%
    \begin{tabular}{|p{1.14\linewidth}|}
    \hline
    \cellcolor{gray1!1}   
    ...\newline\# (Rule6): Harry is kind.\newline\# (Rule7): Harry is smart.\newline\# (Rule8): If someone is kind and big then they are nice.\newline\# (Rule9): If Gary is quiet and Gary is big then Gary is smart.\newline\# (Rule10): If someone is cold then they are green.\newline\# (Rule11): If Harry is smart and Harry is cold then Harry is green.\newline\# (Rule12): Quiet people are big.\newline\# (Rule13): If someone is nice and big then they are cold.\newline\# (Rule14): All nice, quiet people are cold.\newline\# (Rule15): If Gary is nice then Gary is kind.\newline\# (Rule16): Smart people are quiet.\newline\# (Question): Based on the above information, is the following statement true, false, or unknown? Harry is nice.\newline
    \highlightspan{\# (Answer): Start from the object and their condition mentioned in the question to collect relevant facts: Harry, is nice\newline\# KB = \{\} \newline=> Rule6 = `Harry is kind`\newline=> Rule7 = `Harry is smart`\newline\# KB = \{Harry is kind, Harry is smart\} \newline=> F(KB['Harry is smart'], Rule16) => `Harry is quiet`\newline\# KB = \{Harry is kind, Harry is smart, Harry is quiet\} \newline=> F(KB['Harry is quiet'], Rule12) => `Harry is big`\newline\# KB = \{Harry is kind, Harry is smart, Harry is quiet, Harry is big\} \newline=> F(KB['Harry is kind', 'Harry is big'], Rule8) => `Harry is nice`\newline\# KB = \{Harry is kind, Harry is smart, Harry is quiet, Harry is big, Harry is nice\} \newline\# valid the question with current infered premies\newline=> Validate(Question=`Harry is nice`, KB('Harry is nice')) = True. }
    \\
        \hline
    \end{tabular} 
} 
\end{table}

\section{Attention Head Pattern\label{apd_attention_head_pattern}}
\begin{figure*}[!htbp]
    \centering 
    \includegraphics[width=0.99\linewidth, keepaspectratio, 
            trim={2.2cm 2cm 4.8cm 2cm  }, page=1, clip=true]{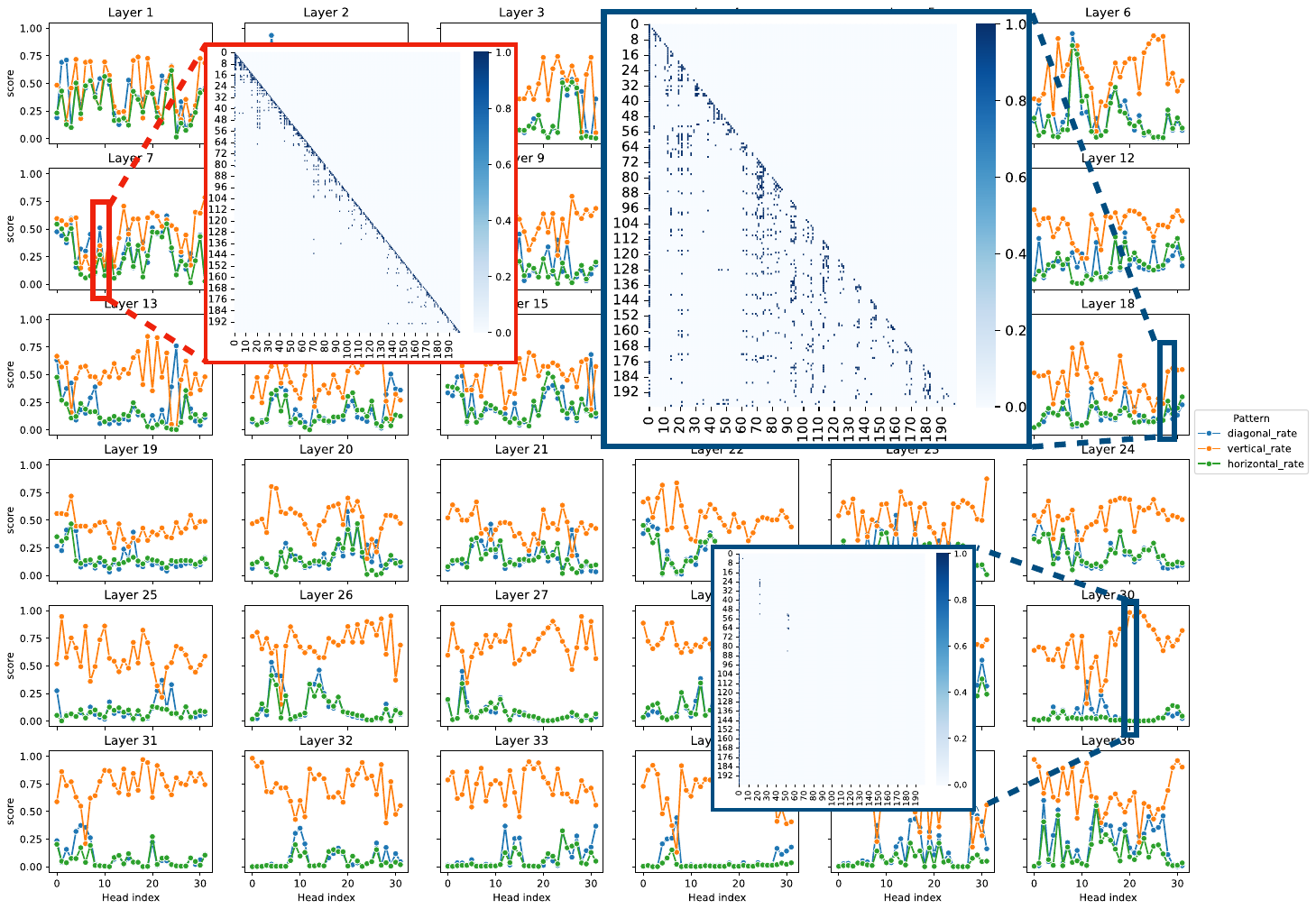}
   
    \caption{Visualization of the distribution of Attention Head Patterns in the \texttt{Qwen3-8B} LLM. The blue, orange, and green lines respectively represent the three scores--diagonal, vertical, and horizontal--computed by the pattern function defined in Eq.~\ref{eq_head_pattern}.
    }\label{fig_attn_pattern_full}
\end{figure*}

In Fig.~\ref{fig_attn_pattern_full}, we present the attention head pattern scores for all heads across 36 layers (each consisting of 32 heads) in the \texttt{Qwen3-8B} LLM. Each attention head is evaluated using three metrics: diagonal, vertical, and horizontal scores. The red zoomed-in region on the left highlights a head with a high diagonal score. The top-right blue zoomed-in region illustrates heads exhibiting high values across all three scores—diagonal, vertical, and horizontal. In contrast, the bottom blue zoomed-in region shows heads characterized by a high vertical score but relatively low horizontal and diagonal scores. 

\section{Prompting Template\label{apd_prompting}} 
The contents of \textit{Symbolic-Aided CoT} prompting \cite{nguyen2025noniterative}  and the \textit{Compact Symbolic-Aided CoT} for the four datasets -- ProofWriter, LogicalDeduction, ProntoQA, FOLIO, and GSM8k -- are presented in Tables~\ref{tab_prompting_PW},~\ref{tab_prompting_LD},~\ref{tab_prompting_Pron},~\ref{tab_prompting_FOLIO}, and Tables~\ref{tab_compact_prompting_PW}, \ref{tab_compact_prompting_LD}, \ref{tab_compact_prompting_pronto}, \ref{tab_compact_prompting_FOLIO}, \ref{tab_compact_prompting_GSM8k}, respectively.

\begin{table}[t]
\small
\centering
\caption{\textit{Symbolic-Aided CoT} prompting for ProofWriter dataset.
\label{tab_prompting_PW} 
}
\resizebox{\linewidth}{!}{%
    \begin{tabular}{|p{1.34\linewidth}|}
    \hline
    \cellcolor{gray1!1}  
  \#\#\# Let us define F as a function that infers new premises based on a given list of facts and rules. Using these facts and rules, provide a reasoning path that leads to one of the values of a Validate function: True, False, or Uncertain.\newline-------\newline\#\#\# Example1: Given list of facts and rules:\newline\# (Rule1): The cow is blue.\newline\# (Rule2): The cow is round.\newline\# (Rule3): The cow likes the lion.\newline\# (Rule4): The cow visits the tiger.\newline\# (Rule5): The lion is cold.\newline\# (Rule6): The lion is nice.\newline\# (Rule7): The lion likes the squirrel.\newline\# (Rule8): The squirrel is round.\newline\# (Rule9): The squirrel sees the lion.\newline\# (Rule10): The squirrel visits the cow.\newline\# (Rule11): The tiger likes the cow.\newline\# (Rule12): The tiger likes the squirrel.\newline\# (Rule13): If something is cold then it visits the tiger.\newline\# (Rule14): If something visits the tiger then it is nice.\newline\# (Rule15): If something sees the tiger and it is young then it is blue.\newline\# (Rule16): If something is nice then it sees the tiger.\newline\# (Rule17): If something likes the squirrel and it likes the cow then it visits the tiger.\newline\# (Rule18): If something is nice and it sees the tiger then it is young.\newline\# (Rule19): If the cow is cold and the cow visits the lion then the lion sees the squirrel.\newline\# (Question): Based on the above information, is the following statement true, false, or uncertain? The tiger is not young.\newline
  \highlightspan{\# (Answer): Start from the object and their condition mentioned in the question to collect relevant facts: The tiger\newline\# KB = \{\} \newline=> Rule11 = `The tiger likes the cow`\newline=> Rule12 = `The tiger likes the squirrel`\newline\# KB = \{The tiger likes the cow, The tiger likes the squirrel\} \newline=> F(KB('The tiger likes the cow', 'The tiger likes the squirrel'), Rule17) => `The tiger visits the tiger`\newline\# KB = \{The tiger likes the cow, The tiger likes the squirrel, The tiger visits the tiger\} \newline=> F(KB['The tiger visits the tiger'], Rule14) => `The tiger is nice`\newline\# KB = \{The tiger likes the cow, The tiger likes the squirrel, The tiger visits the tiger, The tiger is nice\} \newline=> F(KB['The tiger is nice'], Rule16) => `The tiger sees the tiger`\newline\# KB = \{The tiger likes the cow, The tiger likes the squirrel, The tiger visits the tiger, The tiger is nice, The tiger sees the tiger\} \newline=> F(KB('The tiger is nice', 'The tiger sees the tiger'), Rule18) => `The tiger is young`\newline\# KB = \{The tiger likes the cow, The tiger likes the squirrel, The tiger visits the tiger, The tiger is nice, The tiger sees the tiger, The tiger is young\} \newline\# valid the question with current infered premies\newline=> Validate(Question=`The tiger is not young`, KB('The tiger is young')) = False.}
  \newline-------\newline\#\#\# Example2: Given list of facts and rules:\newline\# (Rule1): The dog sees the rabbit.\newline\# (Rule2): The dog sees the squirrel.\newline\# (Rule3): The dog sees the tiger.\newline\# (Rule4): The rabbit eats the dog.\newline\# (Rule5): The rabbit does not eat the tiger.\newline\# (Rule6): The rabbit does not like the tiger.\newline\# (Rule7): The squirrel does not see the rabbit.\newline\# (Rule8): The tiger does not eat the rabbit.\newline\# (Rule9): The tiger is not kind.\newline\# (Rule10): The tiger likes the dog.\newline\# (Rule11): The tiger sees the dog.\newline\# (Rule12): If something is cold then it likes the rabbit.\newline\# (Rule13): If something eats the tiger and it is nice then it likes the rabbit.\newline\# (Rule14): If something likes the squirrel then the squirrel likes the rabbit.\newline\# (Rule15): If something likes the rabbit and the rabbit is kind then it sees the tiger.\newline\# (Rule16): If something likes the tiger then the tiger is young.\newline\# (Rule17): If something is young and it eats the rabbit then it likes the tiger.\newline\# (Rule18): If something sees the rabbit then the rabbit is cold.\newline\# (Rule19): If something likes the rabbit then it likes the squirrel.\newline\# (Rule20): If something likes the squirrel then the squirrel is cold.\newline\# (Question): Based on the above information, is the following statement true, false, or uncertain? The rabbit is cold.\newline
  \highlightspan{\# (Answer): Start from the object and their condition mentioned in the question to collect relevant facts: The rabbit, is cold\newline\# KB = \{\} \newline=> Rule4 = `The rabbit eats the dog`\newline=> Rule5 = `The rabbit does not eat the tiger`\newline=> Rule6 = `The rabbit does not like the tiger`\newline\# KB = \{The rabbit eats the dog, The rabbit does not eat the tiger, The rabbit does not like the tiger\} \newline=> F(KB(Rule1), Rule18) => `The rabbit is cold`\newline\# KB = \{eats the dog, not eat the tiger, not like the tiger, The rabbit is cold\} \newline\# valid the question with current infered premies\newline=> Validate(Question=`The rabbit is cold`, KB('The rabbit is cold')) = True.}
  \newline-------\newline\#\#\# Example3: Given list of facts and rules:\newline\{\{RULE CONTENT\}\}\newline\# (Question): \{\{QUESTION\}\}\newline\# (Answer): 
    \\
        \hline
    \end{tabular} 
} 
\end{table}

\begin{table}[t]
\small
\centering
\caption{\textit{Symbolic-Aided CoT} prompting for LogicalDeduction dataset.
\label{tab_prompting_LD} 
}
\resizebox{\linewidth}{!}{%
    \begin{tabular}{|p{1.31\linewidth}|}
    \hline
    \cellcolor{gray1!1}  
    \#\#\# Let us define F as a function that infers new premises based on a given list of facts and rules. Using these facts and rules, provide a reasoning path that leads to the answer of the question.\newline-------\newline\#\#\# Example1: Given list of rules:\newline\# (Rule1): The following paragraphs each describe a set of seven objects arranged in a fixed order. \newline\# (Rule2): The statements are logically consistent within each paragraph.\newline\# (Rule3): On a branch, there are seven birds: a falcon, an owl, a quail, a raven, a hummingbird, a blue jay, and a crow. \newline\# (Rule4): The quail is to the left of the owl. \newline\# (Rule5): The owl is the second from the left. \newline\# (Rule6): The crow is to the right of the falcon. \newline\# (Rule7): The blue jay is to the right of the hummingbird. \newline\# (Rule8): The hummingbird is to the right of the raven. \newline\# (Rule9): The blue jay is to the left of the falcon.\newline\# (Question):  Which of the following is true?\newline\# (Options):\newline A) The falcon is the second from the left.\newline B) The owl is the second from the left.\newline C) The quail is the second from the left.\newline D) The raven is the second from the left.\newline E) The hummingbird is the second from the left.\newline F) The blue jay is the second from the left.\newline G) The crow is the second from the left.\newline
    \highlightspan{\# (Answer): Since the options ask which object is the second from the left, the objects are ordered from the left:\newline\# KB = []\newline=> F(`ordered from the left`,Rule1) => KB = [1., 2., 3., 4., 5., 6., 7.]\newline=> F(`ordered from the left`,Rule5) => absolute position of second from the left = 2 =>  KB = [1., 2.owl, 3., 4., 5., 6., 7.]\newline=> F(`ordered from the left`,Rule4) => Draft relative position: quail -\textgreater owl\newline=> F(`ordered from the left`,Rule8) => Draft relative position: raven -\textgreater hummingbird => Combine bridge relative: raven -\textgreater hummingbird; quail -\textgreater owl\newline=> F(`ordered from the left`,Rule7) => Draft relative position: hummingbird -\textgreater blue jay => Combine bridge relative: raven -\textgreater hummingbird -\textgreater blue jay; quail -\textgreater owl\newline=> F(`ordered from the left`,Rule9) => Draft relative position: blue jay -\textgreater falcon => Combine bridge relative: raven -\textgreater hummingbird -\textgreater blue jay -\textgreater falcon; quail -\textgreater owl \newline=> F(`ordered from the left`,Rule6) => Draft relative position: falcon -\textgreater crow => Combine bridge relative: raven -\textgreater hummingbird -\textgreater blue jay -\textgreater falcon -\textgreater crow; quail -\textgreater owl\newline=> Combine (KB=[1., 2.owl, 3., 4., 5., 6., 7.], Draft) => with respect to absolute positions in KB: [1.quail, 2.owl, 3.raven, 4.hummingbird, 5.blue jay, 6.falcon, 7.crow]\newline\# Since the options ask which object is the second from the left => absolute position is 2, the correct answer is `owl`. Check all the above options => answer: B}
    \newline-------\newline\#\#\# Example2: Given list of rules:\newline\# (Rule1): The following paragraphs each describe a set of five objects arranged in a fixed order. \newline\# (Rule2): The statements are logically consistent within each paragraph.\newline\# (Rule3): A fruit stand sells five fruits: watermelons, apples, kiwis, cantaloupes, and mangoes. \newline\# (Rule4): The cantaloupes are the second-cheapest. \newline\# (Rule5): The mangoes are more expensive than the watermelons. \newline\# (Rule6): The kiwis are less expensive than the apples. \newline\# (Rule7): The kiwis are more expensive than the mangoes.\newline\# (Question):  Which of the following is true?\newline\# (Options):\newline A) The watermelons are the cheapest.\newline B) The apples are the cheapest.\newline C) The kiwis are the cheapest.\newline D) The cantaloupes are the cheapest.\newline E) The mangoes are the cheapest.\newline
    \highlightspan{\# (Answer): Since the options ask which fruit is the cheapest, the objects are ordered from cheapest:\newline\# KB = []\newline=> F(`ordered from cheapest`,Rule1) => KB = [1., 2., 3., 4., 5.]\newline=> F(`ordered from cheapest`,Rule4) => absolute position of second-cheapest = 2 => KB = [ 1., 2.cantaloupes, 3., 4., 5.]\newline=> F(`ordered from cheapest`,Rule5) => Draft relative position: watermelons -\textgreater mangoes\newline=> F(`ordered from cheapest`,Rule7) => Draft relative position: mangoes -\textgreater kiwis => Combine bridge relative: watermelons -\textgreater mangoes -\textgreater kiwis \newline=> F(`ordered from cheapest`,Rule6) => Draft relative position: kiwis -\textgreater apples => Combine bridge relative: watermelons -\textgreater mangoes -\textgreater kiwis -\textgreater apples\newline=> Combine (KB=[ 1., 2.cantaloupes, 3., 4., 5.], Draft) => with respect to absolute positions in KB:  [1.watermelons, 2.cantaloupes, 3.mangoes, 4.kiwis, 5.apples]\newline\# Since the options ask which fruit is the cheapest => absolute position is 1, the correct answer is `watermelons`. Check all the above options => answer: A}
    \newline-------\newline\#\#\# Example3: Given list of rules:\newline\{\{RULE CONTENT\}\}\newline\# (Question): \{\{QUESTION\}\}\newline\# (Options): \{\{OPTIONS\}\}\newline\# (Answer): Since the options ask 
    \\
        \hline
    \end{tabular} 
} 
\end{table}

\begin{table}[!t]
\small
\centering
\caption{\textit{Symbolic-Aided CoT} prompting for ProntoQA dataset.
\label{tab_prompting_Pron} 
}
\resizebox{\linewidth}{!}{%
    \begin{tabular}{|p{1.55\linewidth}|}
    \hline
    \cellcolor{gray1!1}  
\#\#\# Let us define F as a function that determines whether a statement is True or False, based on a given list of facts and rules. Using these facts and rules, provide a reasoning path that leads to one of the values: `True` or `False`.\newline-------\newline\#\#\# Example1: Given list of facts and rules: \newline\# (Rule1): If K then P; \newline\# (Rule2): Q is true; \newline\# (Rule3): If Q then K. \newline\# (Question): truth value of P? \newline
\highlightspan{\# (Answer): Start from the object mentioned in the question: P\newline\# KB = \{Q\} \newline=> F(KB['Q'], Rule3) => `K`\newline\# KB = \{Q, K\}\newline=> F(KB['K'], Rule1) => `P`\newline\# KB = \{Q, K, P\}\newline=> Validate(KB, Question=`P`) = True.}
\newline-------\newline\#\#\# Example2: Given list of facts and rules:\newline\# (Rule1): Jompuses are not shy. \newline\# (Rule2): Jompuses are yumpuses. \newline\# (Rule3): Zumpuses are rompuses. \newline\# (Rule4): Each yumpus is a dumpus. \newline\# (Rule5): Dumpuses are not wooden. \newline\# (Rule6): Dumpuses are wumpuses. \newline\# (Rule7): Wumpuses are red. \newline\# (Rule8): Every wumpus is an impus. \newline\# (Rule9): Each impus is opaque. \newline\# (Rule10): Impuses are tumpuses. \newline\# (Rule11): Numpuses are sour. \newline\# (Rule12): Tumpuses are not sour. \newline\# (Rule13): Tumpuses are vumpuses. \newline\# (Rule14): Vumpuses are earthy. \newline\# (Rule15): Every vumpus is a zumpus. \newline\# (Rule16): Zumpuses are small. \newline\# (Rule17): Each yumpus is aggressive. \newline\# (Rule18): Max is a yumpus.\newline\# (Question): Is the following statement true or false? Max is sour.\newline
\highlightspan{\# (Answer): Start from the object mentioned in the question: Max\newline\# KB = \{\} \newline=> F(KB, Rule17) => `Max is a yumpus`\newline\# KB = \{yumpus\} \newline=> F(KB['yumpus'], Rule17) => `Max is aggressive`\newline=> F(KB['yumpus'], Rule4) => `Max is dumpus`\newline\# KB = \{yumpus, aggressive, dumpus\} \newline=> F(KB['dumpus'], Rule5) => `Max is not wooden`\newline=> F(KB['dumpus'], Rule6) => `Max is wumpus`\newline\# KB = \{yumpus, aggressive, dumpus, not wooden, wumpus\}  \newline=> F(KB['wumpus'], Rule7) => `Max is an red`\newline=> F(KB['wumpus'], Rule8) => `Max is an impus`\newline\# KB = \{yumpus, aggressive, dumpus, not wooden, wumpus, red, impus\}  \newline=> F(KB['impus'], Rule9) => `Max is a opaque`\newline=> F(KB['impus'], Rule10) => `Max is a tumpus`\newline\# KB = \{yumpus, aggressive, dumpus, not wooden, wumpus, red, impus, opaque, tumpus\}  \newline=> F(KB['tumpus'], Rule12) => `Max is a not sour`\newline=> F(KB['tumpus'], Rule13) => `Max is a vumpus`\newline\# KB = \{yumpus, aggressive, dumpus, not wooden, wumpus, red, impus, opaque, tumpus, not sour, vumpus\}  \newline\# valid the question with current infered premies\newline=> Validate(KB['not sour'], Question=`Max is sour`) = False.}
\newline-------\newline\#\#\# Example3: Given list of facts and rules:\newline\# (Rule1): Every tumpus is not angry. \newline\# (Rule2): Tumpuses are rompuses. \newline\# (Rule3): Every numpus is not bright. \newline\# (Rule4): Rompuses are not luminous. \newline\# (Rule5): Rompuses are yumpuses. \newline\# (Rule6): Yumpuses are transparent. \newline\# (Rule7): Yumpuses are zumpuses. \newline\# (Rule8): Each zumpus is not bitter. \newline\# (Rule9): Zumpuses are impuses. \newline\# (Rule10): Impuses are red. \newline\# (Rule11): Each impus is a dumpus. \newline\# (Rule12): Every dumpus is happy. \newline\# (Rule13): Each dumpus is a vumpus. \newline\# (Rule14): Vumpuses are bright. \newline\# (Rule15): Every vumpus is a jompus. \newline\# (Rule16): Jompuses are large. \newline\# (Rule17): Each jompus is a wumpus. \newline\# (Rule18): Stella is a yumpus.\newline\# (Question): Is the following statement true or false? Stella is bright.\newline
\highlightspan{\# (Answer): Start from the object mentioned in the question: Stella\newline\# KB = \{\} \newline=> F(KB, Rule17) => `Stella is a yumpus`\newline\# KB = \{yumpus\} \newline=> F(KB['yumpus'], Rule6) => `Stella is transparent`\newline=> F(KB['yumpus'], Rule7) => `Stella is zumpus`\newline\# KB = \{yumpus, transparent, zumpus\} \newline=> F(KB['zumpus'], Rule9) => `Stella is impus` \newline=> F(KB['zumpus'], Rule8) => `Stella is not bitter`\newline\# KB = \{yumpus, transparent, zumpus, impus, not bitter\} \newline=> F(KB['impus'], Rule10) => `Stella is red`\newline=> F(KB['impus'], Rule11) => `Stella is dumpus` \newline\# KB = \{yumpus, transparent, zumpus, not bitter, impus, red, dumpus\} \newline=> F(KB['dumpus'], Rule13) => `Stella is a vumpus`\newline=> F(KB['dumpus'], Rule12) => `Stella is happy` \newline\# KB = \{yumpus, transparent, zumpus, not bitter, impus, red, dumpus, vumpus, happy\} \newline=> F(KB['vumpus'], Rule14) => `Stella is bright`\newline\# KB = \{yumpus, transparent, zumpus, not bitter, impus, red, dumpus, vumpus, happy, bright\} \newline\# valid the question with current infered premies\newline=> Validate(KB['bright'], Question=`Stella is bright`) = True.}
\newline-------\newline\#\#\# Example4: Given list of facts and rules:\newline\texttt{\{\{RULE CONTENT\}\}}\newline\# (Question): \texttt{\{\{QUESTION\}\}}\newline\# (Answer):
        \\
        \hline
    \end{tabular} 
} 
\end{table}

\begin{table*}[!t]
\small
\centering
\caption{\textit{Symbolic-Aided CoT} prompting for FOLIO dataset.
\label{tab_prompting_FOLIO} 
}
\resizebox{\linewidth}{!}{%
    \begin{tabular}{|p{1.2\linewidth}|}
    \hline
    \cellcolor{gray1!1}  
    \#\#\# Let us define F as a function that infers new premises based on a given list of facts and rules. Using these facts and rules, provide a reasoning path that leads to one of the values of a Validate function: True, False, or Uncertain. (Guidelines: True – If the hypothesis can be logically proven correct using only the provided facts and rules. False – If the hypothesis can be logically disproven (the facts or rules directly contradict the hypothesis). Uncertain – If there is not enough information to prove or disprove the hypothesis. Important Rules: Only use the provided facts and rules as your knowledge base (KB); Do not use outside knowledge or make assumptions not supported by the KB; Avoid circular or infinite reasoning. Each rule may only be applied once to prevent loops; If the reasoning process leads to conflicting conclusions, classify as Uncertain. Conditions in if-then rules are not confirmed facts. When rules reference vague entities (\textit{e.g.,} ‘an object exists’ or ‘some objects’), the outcome is often uncertain. Output Format: Return only one word: True, False, or Uncertain)\newline-------\newline\#\#\# Example1: Given list of facts and rules:\newline\# (Rule1): The Blake McFall Company Building is a commercial warehouse listed on the National Register of Historic Places.\newline\# (Rule2): The Blake McFall Company Building was added to the National Register of Historic Places in 1990.\newline\# (Rule3): The Emmet Building is a five-story building in Portland, Oregon.\newline\# (Rule4): The Emmet Building was built in 1915. \newline\# (Rule5): The Emmet Building is another name for the Blake McFall Company Building. \newline\# (Rule6): John works at the Emmet Building.\newline\# (Question): Based on the above information, is the following statement true, false, or uncertain? The Blake McFall Company Building is located in Portland, Oregon.\newline
    \highlightspan{\# (Answer): Start with the object and its condition mentioned in the question to collect relevant facts. In this case, the main object is `Blake McFall Company Building`. To verify the statement — `The Blake McFall Company Building is located in Portland, Oregon` - is True, False, or Uncertain (Contradiction or not enough information to conclusion).\newline=> KB = \{\}\newline\# collect all the facts or assumption conditions \newline=> Rule1 = KB(`The Blake McFall Company Building is a commercial warehouse listed on the National Register of Historic Places`)\newline=> Rule2 = KB(`The Blake McFall Company Building was added to the National Register of Historic Places in 1990`)\newline=> Rule5 = KB(`The Emmet Building is another name for the Blake McFall Company Building`)\newline=> F(KB(`The Emmet Building is another name for the Blake McFall Company Building`), Rule3) => KB(`The Blake McFall Company Building is a five-story building in Portland, Oregon`)\newline\# valid the question with current infered premies\newline=> Validate(Question=`The Blake McFall Company Building is located in Portland, Oregon`, KB(`The Blake McFall Company Building is a five-story building in Portland, Oregon`)) = True.}
    \newline-------\newline\#\#\# Example2: Given list of facts and rules:\newline\# (Rule1): People eat meat regularly or are vegetation. \newline\# (Rule2): If people eat meat regularly, then they enjoy eating hamburgers and steaks.\newline\# (Rule3): All people who are vegetarian are conscious of the environment or their health. \newline\# (Rule4): If people are conscious about the environment or their health, then they do not go to fast food places often.\newline\# (Rule5): If people have busy schedules without time to cook, then they go to fast food places often. \newline\# (Rule6): If Jeremy does not both go to fast food places often and is conscious about the environment or their health, then he goes to fast food places often.\newline\# (Question): Based on the above information, is the following statement true, false, or uncertain?  If Jeremy has a busy schedule without time to cook, then Jeremy does not enjoy eating hamburgers and steaks.\newline
    \highlightspan{\# (Answer): Start with the object and its condition mentioned in the question to collect relevant facts. In this case, the main object is `Jeremy`. To verify the assumption — if `Jeremy has a busy schedule without time to cook` — determine whether the conclusion `Jeremy does not enjoy eating hamburgers and steaks` is True, False, or Uncertain (Contradiction or not enough information to conclusion).\newline=> KB = \{\}\newline\# collect all the facts or assumption conditions \newline=> KB (`Jeremy has a busy schedule without time to cook`)\newline=> F(KB(`Jeremy has a busy schedule without time to cook`), Rule5) => KB(`Jeremy goes to fast food places often`)\newline=> F(KB(`Jeremy goes to fast food places often`), Rule4) => KB(`Jeremy is not conscious about the environment or his health`)\newline=> F(KB(`Jeremy is not conscious about the environment or his health`), Rule3) => KB(`Jeremy is not vegetarian`)\newline=> F(KB(`Jeremy is not vegetarian`), Rule1) => KB(`Jeremy eats meat regularly`)\newline=> F(KB(`Jeremy eats meat regularly`), Rule2) => KB(`Jeremy enjoys eating hamburgers and steaks`)\newline\# valid the question with current infered premies\newline=> Validate(Question=`If Jeremy has a busy schedule without time to cook, then Jeremy does not enjoy eating hamburgers and steaks`, KB(`Jeremy enjoys eating hamburgers and steaks`)) = False.}
    \newline-------\newline\#\#\# Example3: Given list of facts and rules:
    \newline\# (Rule1): Pasifika New Zealanders are a pan-ethnic group of New Zealanders.\newline\# (Rule2): Asian New Zealanders are a pan-ethnic group of New Zealanders.\newline\# (Rule3): Pasifika New Zealanders are not Asian New Zealanders.\newline\# (Rule4): Pasifika New Zealanders speak Samoan.\newline\# (Rule5): Joe is a Pasifika New Zealander.\newline
    \# (Rule6): Amy speaks Samoan.\newline\# (Question): Based on the above information, is the following statement true, false, or uncertain? Amy is an Pasifika New Zealander.\newline
    \highlightspan{\# (Answer): Start with the object and its condition mentioned in the question to collect relevant facts. In this case, the main object is `Amy`. To verify the statement — `Amy is an Pasifika New Zealander` - is True, False, or Uncertain (Contradiction or not enough information to conclusion).
    \newline=> KB = \{\}\newline\# collect all the facts or assumption conditions \newline=> KB (`Amy speaks Samoan`)\newline=> F(KB(`Amy speaks Samoan`), Rule4) => Uncertain(`Amy is a Pasifika New Zealander`)\newline\# valid the question with current infered premies\newline=> Validate(Question=`Amy is a Pasifika New Zealander`, KB) = Uncertain.}
    \newline------- \newline\#\#\# Example4: Given list of facts and rules:\newline\texttt{\{\{RULE CONTENT\}\}}\newline\# (Question):  Based on the above information, is the following statement true, false, or uncertain? \texttt{\{\{QUESTION\}\}}\newline\# (Answer):
        \\
        \hline
    \end{tabular} 
} 
\end{table*}

\begin{table*}[!t]
\small
\centering
\caption{\textit{Compact Symbolic-Aided CoT} prompting for ProofWriter dataset.
\label{tab_compact_prompting_PW} 
}
\resizebox{\linewidth}{!}{%
    \begin{tabular}{|p{1.\linewidth}|}
    \hline
    \cellcolor{gray1!1}  
    \#\#\# Let us define F as a function that determines whether a statement is True, False, or Uncertain, based on a given list of facts and rules. Using these facts and rules, provide a reasoning path that leads to one of the values: `True`, `False`, or `Uncertain`.\newline-------\newline\#\#\# Example1: Given list of facts and rules: \newline\# (Rule1): If K then P; \newline\# (Rule2): Q is true; \newline\# (Rule3): If Q then K. \newline\# (Question): truth value of P? \newline
    \highlightspan{\# (Answer): F(Rule2) => F(Rule2, Rule3, Output1=`K`) => F(Output1, Rule1, Output2=`P`) => F(Question) = True.}
    \newline-------\newline\#\#\# Example2: Given list of facts and rules:\newline\# (Rule1): The cow is blue.\newline\# (Rule2): The cow is round.\newline\# (Rule3): The cow likes the lion.\newline\# (Rule4): The cow visits the tiger.\newline\# (Rule5): The lion is cold.\newline\# (Rule6): The lion is nice.\newline\# (Rule7): The lion likes the squirrel.\newline\# (Rule8): The squirrel is round.\newline\# (Rule9): The squirrel sees the lion.\newline\# (Rule10): The squirrel visits the cow.\newline\# (Rule11): The tiger likes the cow.\newline\# (Rule12): The tiger likes the squirrel.\newline\# (Rule13): If something is cold then it visits the tiger.\newline\# (Rule14): If something visits the tiger then it is nice.\newline\# (Rule15): If something sees the tiger and it is young then it is blue.\newline\# (Rule16): If something is nice then it sees the tiger.\newline\# (Rule17): If something likes the squirrel and it likes the cow then it visits the tiger.\newline\# (Rule18): If something is nice and it sees the tiger then it is young.\newline\# (Rule19): If the cow is cold and the cow visits the lion then the lion sees the squirrel.\newline\# (Question): Based on the above information, is the following statement true, false, or uncertain? The tiger is not young.\newline
    \highlightspan{\# (Answer): F((Rule11, Rule12), Rule17, Output1=`The tiger visits the tiger`) => F(Output1, Rule14, Output2=`The tiger is nice`) => F(Output2, Rule18, Output=`The tiger is young`) => F(Question) = False.}
    \newline-------\newline\#\#\# Example3: Given list of facts and rules:\newline\# (Rule1): The dog sees the rabbit.\newline\# (Rule2): The dog sees the squirrel.\newline\# (Rule3): The dog sees the tiger.\newline\# (Rule4): The rabbit eats the dog.\newline\# (Rule5): The rabbit does not eat the tiger.\newline\# (Rule6): The rabbit does not like the tiger.\newline\# (Rule7): The squirrel does not see the rabbit.\newline\# (Rule8): The tiger does not eat the rabbit.\newline\# (Rule9): The tiger is not kind.\newline\# (Rule10): The tiger likes the dog.\newline\# (Rule11): The tiger sees the dog.\newline\# (Rule12): If something is cold then it likes the rabbit.\newline\# (Rule13): If something eats the tiger and it is nice then it likes the rabbit.\newline\# (Rule14): If something likes the squirrel then the squirrel likes the rabbit.\newline\# (Rule15): If something likes the rabbit and the rabbit is kind then it sees the tiger.\newline\# (Rule16): If something likes the tiger then the tiger is young.\newline\# (Rule17): If something is young and it eats the rabbit then it likes the tiger.\newline\# (Rule18): If something sees the rabbit then the rabbit is cold.\newline\# (Rule19): If something likes the rabbit then it likes the squirrel.\newline\# (Rule20): If something likes the squirrel then the squirrel is cold.\newline\# (Question): Based on the above information, is the following statement true, false, or uncertain? The rabbit is cold.\newline
    \highlightspan{\# (Answer): F(Rule1, Rule18, Output1=`The rabbit is cold`) => F(Question) = True. }
  \newline-------\newline\#\#\# Example3: Given list of facts and rules:\newline\{\{RULE CONTENT\}\}\newline\# (Question): Based on the above information, is the following statement true, false, or unknown?  \{\{QUESTION\}\}\newline\# (Answer): 
    \\
        \hline
    \end{tabular} 
} 
\end{table*}


\begin{table*}[!t]
\small
\centering
\caption{\textit{Compact Symbolic-Aided CoT} prompting for LogicalDeduction dataset.
\label{tab_compact_prompting_LD} 
}
\resizebox{\linewidth}{!}{%
    \begin{tabular}{|p{1.\linewidth}|}
    \hline
    \cellcolor{gray1!1}  
    \#\#\# Let us define F as a function that infers new premises based on a given list of facts and rules. Using these facts and rules, provide a reasoning path that leads to the answer of the question.\newline-------\newline\#\#\# Example1: Given list of rules:\newline\# (Rule1): The following paragraphs each describe a set of seven objects arranged in a fixed order. \newline\# (Rule2): The statements are logically consistent within each paragraph.\newline\# (Rule3): On a branch, there are seven birds: a falcon, an owl, a quail, a raven, a hummingbird, a blue jay, and a crow. \newline\# (Rule4): The quail is to the left of the owl. \newline\# (Rule5): The owl is the second from the left. \newline\# (Rule6): The crow is to the right of the falcon. \newline\# (Rule7): The blue jay is to the right of the hummingbird. \newline\# (Rule8): The hummingbird is to the right of the raven. \newline\# (Rule9): The blue jay is to the left of the falcon.\newline\# (Question):  Which of the following is true?\newline\# (Options):\newline A) The falcon is the second from the left.\newline B) The owl is the second from the left.\newline C) The quail is the second from the left.\newline D) The raven is the second from the left.\newline E) The hummingbird is the second from the left.\newline F) The blue jay is the second from the left.\newline G) The crow is the second from the left.\newline
    \highlightspan{\# (Answer): Since the options ask which object is the second from the left, the objects are ordered from the left:\newline\# => F(Rule1, Output=`[1., 2., 3., 4., 5., 6., 7.]`) => F(Rule5, Output=`[1., 2.owl, 3., 4., 5., 6., 7.]`) => F(Rule4, Output=`quail -> owl`) => F(Rule8, Output=`raven -> hummingbird; quail -> owl`) => F(Rule7, Output=`raven -> hummingbird -> blue jay; quail -> owl`) => F(Rule9, Output=`raven -> hummingbird -> blue jay -> falcon; quail -> owl`) => F(Rule6, Output=`raven -> hummingbird -> blue jay -> falcon -> crow; quail -> owl`) => Combination(Output=`[1.quail, 2.owl, 3.raven, 4.hummingbird, 5.blue jay, 6.falcon, 7.crow]`) => Validate(Question=`which object is the second from the left`, Output=`owl`) => answer: B}
    \newline-------\newline\#\#\# Example2: Given list of rules:\newline\# (Rule1): The following paragraphs each describe a set of five objects arranged in a fixed order. \newline\# (Rule2): The statements are logically consistent within each paragraph.\newline\# (Rule3): A fruit stand sells five fruits: watermelons, apples, kiwis, cantaloupes, and mangoes. \newline\# (Rule4): The cantaloupes are the second-cheapest. \newline\# (Rule5): The mangoes are more expensive than the watermelons. \newline\# (Rule6): The kiwis are less expensive than the apples. \newline\# (Rule7): The kiwis are more expensive than the mangoes.\newline\# (Question):  Which of the following is true?\newline\# (Options):\newline A) The watermelons are the cheapest.\newline B) The apples are the cheapest.\newline C) The kiwis are the cheapest.\newline D) The cantaloupes are the cheapest.\newline E) The mangoes are the cheapest.\newline
    \highlightspan{\# (Answer): Since the options ask which fruit is the cheapest, the objects are ordered from cheapest:\newline\# => F(Rule1, Output=`[1., 2., 3., 4., 5.]`) => F(Rule4, Output=`[ 1., 2.cantaloupes, 3., 4., 5.]`) => F(Rule5, Output=`watermelons -> mangoes`) => F(Rule7, Output=`watermelons -> mangoes -> kiwis`) => F(Rule6, Output=`watermelons -> mangoes -> kiwis -> apples`) => Combination(Output=`[1.watermelons, 2.cantaloupes, 3.mangoes, 4.kiwis, 5.apples]`) => Validate(Question=`which fruit is the cheapest`, Output=`watermelons`) => answer: A}
    \newline-------\newline\#\#\# Example3: Given list of rules:\newline\{\{RULE CONTENT\}\}\newline\# (Question): \{\{QUESTION\}\}\newline\# (Options): \{\{OPTIONS\}\}\newline\# (Answer): Since the options ask 
    \\
        \hline
    \end{tabular} 
} 
\end{table*}

\begin{table*}[!t]
\small
\centering
\caption{\textit{Compact Symbolic-Aided CoT} prompting for ProntoQA dataset.
\label{tab_compact_prompting_pronto} 
}
\resizebox{\linewidth}{!}{%
    \begin{tabular}{|p{1.\linewidth}|}
    \hline
    \cellcolor{gray1!1}  
    \#\#\# Let us define F as a function that determines whether a statement is True or False, based on a given list of facts and rules. Using these facts and rules, provide a reasoning path that leads to one of the values: `True` or `False`.\newline-------\newline\#\#\# Example1: Given list of facts and rules: \newline\# (Rule1): If K then P; \newline\# (Rule2): Q is true; \newline\# (Rule3): If Q then K. \newline\# (Question): truth value of P? \newline
    \highlightspan{ \# (Answer): Start from the object mentioned in the question: P\newline\# F(Rule2) => F(Rule2, Rule3, Output1=`K`) => F(Output1, Rule1, Output2=`P`) => Validate(Output2, Question=`P`) = True.}
    \newline-------\newline\#\#\# Example2: Given list of facts and rules:\newline\# (Rule1): Jompuses are not shy. \newline\# (Rule2): Jompuses are yumpuses. \newline\# (Rule3): Each yumpus is aggressive. \newline\# (Rule4): Each yumpus is a dumpus. \newline\# (Rule5): Dumpuses are not wooden. \newline\# (Rule6): Dumpuses are wumpuses. \newline\# (Rule7): Wumpuses are red. \newline\# (Rule8): Every wumpus is an impus. \newline\# (Rule9): Each impus is opaque. \newline\# (Rule10): Impuses are tumpuses. \newline\# (Rule11): Numpuses are sour. \newline\# (Rule12): Tumpuses are not sour. \newline\# (Rule13): Tumpuses are vumpuses. \newline\# (Rule14): Vumpuses are earthy. \newline\# (Rule15): Every vumpus is a zumpus. \newline\# (Rule16): Zumpuses are small. \newline\# (Rule17): Zumpuses are rompuses. \newline\# (Rule18): Max is a yumpus.\newline\# (Question): Is the following statement true or false? Max is sour.\newline
     \highlightspan{\# (Answer): Start from the object mentioned in the question: Max\newline\# F(Rule18, Rule4, Output1=`Max is dumpus`) => F(Output1, Rule6, Output2=`Max is wumpus`) => F(Output2, Rule8, Output3=`Max is an impus`) => F(Output3, Rule10, Output4=`Max is a tumpus`) => F(Output4, Rule12, Output5=`Max is not sour`) => Validate(Output5, Question=`Max is sour`) = False.}
    \newline-------\newline\#\#\# Example3: Given list of facts and rules:\newline\# (Rule1): Every tumpus is not angry. \newline\# (Rule2): Tumpuses are rompuses. \newline\# (Rule3): Every numpus is not bright. \newline\# (Rule4): Rompuses are not luminous. \newline\# (Rule5): Rompuses are yumpuses. \newline\# (Rule6): Yumpuses are transparent. \newline\# (Rule7): Yumpuses are zumpuses. \newline\# (Rule8): Each zumpus is not bitter. \newline\# (Rule9): Zumpuses are impuses. \newline\# (Rule10): Impuses are red. \newline\# (Rule11): Each impus is a dumpus. \newline\# (Rule12): Every dumpus is happy. \newline\# (Rule13): Each dumpus is a vumpus. \newline\# (Rule14): Vumpuses are bright. \newline\# (Rule15): Every vumpus is a jompus. \newline\# (Rule16): Jompuses are large. \newline\# (Rule17): Each jompus is a wumpus. \newline\# (Rule18): Stella is a yumpus.\newline\# (Question): Is the following statement true or false? Stella is bright.\newline
    \highlightspan{\# (Answer): Start from the object mentioned in the question: Stella\newline\# F(Rule18, Rule7, Output1=`Stella is a zumpus`) => F(Output1, Rule9, Output2=`Stella is an impus`) => F(Output2, Rule11, Output3=`Stella is a dumpus`) => F(Output3, Rule13, Output4=`Stella is a vumpus`) => F(Output4, Rule14, Output5=`Stella is bright`) => Validate(Output5, Question=`Stella is bright`) = True.}
    \newline-------\newline\#\#\# Example3: Given list of rules:\newline\{\{RULE CONTENT\}\}\newline\# (Question): \{\{QUESTION\}\}\newline\# (Answer): 
    \\
        \hline
    \end{tabular} 
} 
\end{table*}

\begin{table*}[!t]
\small
\centering
\caption{\textit{Compact Symbolic-Aided CoT} prompting for FOLIO dataset.
\label{tab_compact_prompting_FOLIO} 
}
\resizebox{\linewidth}{!}{%
    \begin{tabular}{|p{1\linewidth}|}
    \hline
    \cellcolor{gray1!1}  
    \#\#\# Let us define F as a function that infers new premises based on a given list of facts and rules. Using these facts and rules, provide a reasoning path that leads to one of the values of a Validate function: True, False, or Uncertain.\newline-------\newline\#\#\# Example1: Given list of facts and rules:\newline\# (Rule1): The Blake McFall Company Building is a commercial warehouse listed on the National Register of Historic Places.\newline\# (Rule2): The Blake McFall Company Building was added to the National Register of Historic Places in 1990.\newline\# (Rule3): The Emmet Building is a five-story building in Portland, Oregon.\newline\# (Rule4): The Emmet Building was built in 1915. \newline\# (Rule5): The Emmet Building is another name for the Blake McFall Company Building. \newline\# (Rule6): John works at the Emmet Building.\newline\# (Question): Based on the above information, is the following statement true, false, or uncertain? The Blake McFall Company Building is located in Portland, Oregon.\newline
    \highlightspan{\# (Answer): Start with the object and its condition mentioned in the question to collect relevant facts.\newline\# => F(Rule5, Rule3, Output1=`The Blake McFall Company Building is a five-story building in Portland, Oregon`) => Validate(Output1, Question=`The Blake McFall Company Building is located in Portland, Oregon`) = True.}
    \newline-------\newline\#\#\# Example2: Given list of facts and rules:\newline\# (Rule1): People eat meat regularly or are vegetation. \newline\# (Rule2): If people eat meat regularly, then they enjoy eating hamburgers and steaks.\newline\# (Rule3): All people who are vegetarian are conscious of the environment or their health. \newline\# (Rule4): If people are conscious about the environment or their health, then they do not go to fast food places often.\newline\# (Rule5): If people have busy schedules without time to cook, then they go to fast food places often. \newline\# (Rule6): If Jeremy does not both go to fast food places often and is conscious about the environment or their health, then he goes to fast food places often.\newline\# (Question): Based on the above information, is the following statement true, false, or uncertain?  If Jeremy has a busy schedule without time to cook, then Jeremy does not enjoy eating hamburgers and steaks.\newline
    \highlightspan{\# (Answer): Start with the object and its condition mentioned in the question to collect relevant facts.\newline\# => F(Assume(`Jeremy has a busy schedule without time to cook`), Rule5, Output1=`Jeremy goes to fast food places often`) => F(Output1, Rule4, Output2=`Jeremy is not conscious about the environment or his health`) => F(Output2, Rule3, Output3=`Jeremy is not vegetarian`) => F(Output3, Rule1, Output4=`Jeremy eats meat regularly`) => F(Output4, Rule2, Output5=`Jeremy enjoys eating hamburgers and steaks`) => Validate(Output5, Question=`If Jeremy has a busy schedule without time to cook, then Jeremy does not enjoy eating hamburgers and steaks`) = False.}
    \newline-------\newline\#\#\# Example3: Given list of facts and rules:\newline\# (Rule1): Pasifika New Zealanders are a pan-ethnic group of New Zealanders.\newline\# (Rule2): Asian New Zealanders are a pan-ethnic group of New Zealanders.\newline\# (Rule3): Pasifika New Zealanders are not Asian New Zealanders.\newline\# (Rule4): Pasifika New Zealanders speak Samoan.\newline\# (Rule5): Joe is a Pasifika New Zealander.\newline\# (Rule6): Amy speaks Samoan.\newline\# (Question): Based on the above information, is the following statement true, false, or uncertain? Amy is an Pasifika New Zealander.\newline
    \highlightspan{\# (Answer): Start with the object and its condition mentioned in the question to collect relevant facts.\newline\# => F(Rule6, Rule4, Output1=Uncertain(`Amy is a Pasifika New Zealander`) => Validate(Output1, Question=`Amy is a Pasifika New Zealander`) = Uncertain.}
    \newline-------\newline\#\#\# Example4: Given list of facts and rules:\newline\texttt{\{\{RULE CONTENT\}\}}\newline\# (Question): Based on the above information, is the following statement true, false, or uncertain? \texttt{\{\{QUESTION\}\}}\newline\# (Answer):
        \\
        \hline
    \end{tabular} 
} 
\end{table*}

\begin{table*}[!t]
\small
\centering
\caption{\textit{Compact Symbolic-Aided CoT} prompting for GSM8k dataset.
\label{tab_compact_prompting_GSM8k} 
}
\resizebox{\linewidth}{!}{%
    \begin{tabular}{|p{1.\linewidth}|}
    \hline
    \cellcolor{gray1!1}
       \#\#\# Let us define F as a function to infer solution from rules (mathematical conditions). \newline-------\newline Context:  There are 15 trees in the grove. Grove workers will plant trees in the grove today. After they are done, there will be 21 trees. How many trees did the grove workers plant today?\newline
       \highlightspan{\# (Rule1): There are 15 trees in the grove\newline\# (Rule2): Grove workers will plant trees in the grove today. After they are done, there will be 21 trees \newline\# (Question): How many trees did the grove workers plant today?\newline\# (Answer): \newline Variables = \{original\_trees, trees\_after\_some\_more\_were\_planted, trees\_were\_planted\}\newline=> F(Rule1) => original\_trees = 15 \newline=> F(Rule2) => trees\_after\_some\_more\_were\_planted = 21\newline=> F(Question) => trees\_were\_planted = trees\_after\_some\_more\_were\_planted - original\_trees = 21 - 5 = 6.}
       \newline-------\newline Context:  If there are 3 cars in the parking lot and 2 more cars arrive, how many cars are in the parking lot?\newline
       \highlightspan{\# (Rule1): there are 3 cars in the parking lot\newline\# (Rule2): 2 more cars arrive \newline\# (Question): how many cars are in the parking lot?\newline\# (Answer): \newline Variables = \{original\_cars, new\_cars\_arrvie, current\_cars\}\newline=> F(Rule1) => original\_cars = 3\newline=> F(Rule2) => new\_cars\_arrvie = 2\newline=> F(Question) => current\_cars = original\_cars + new\_cars\_arrvie = 5.}
       \newline-------\newline Context:  There were nine computers in the server room. Five more computers were installed each day, from monday to thursday. How many computers are now in the server room?\newline
       \highlightspan{\# (Rule1): There were nine computers in the server room\newline\# (Rule2): Five more computers were installed each day, from monday to thursday\newline\# (Question): How many computers are now in the server room?\newline\# (Answer): \newline Variables = \{original\_computers, computers\_installed\_each\_day, number\_days, number\_computer\_now\}\newline=> F(Rule1) => original\_computers = 9\newline=> F(Rule2) => computers\_installed\_each\_day = 5; number\_days = 4 \# (from monday to thursday) => number\_new\_computers\_installed = 5 * 4 = 20\newline=> F(Question) => number\_computer\_now = original\_computers + number\_new\_computers\_installed = 9 + 20 = 29.}
       \newline-------\newline Context:  Michael had 58 golf balls. On tuesday, he lost 23 golf balls. On wednesday, he lost 2 more. How many golf balls did he have at the end of wednesday?\newline
       \highlightspan{\# (Rule1): Michael had 58 golf balls\newline\# (Rule2): On tuesday, he lost 23 golf balls\newline\# (Rule3): On wednesday, he lost 2 more\newline\# (Question): How many golf balls did he have at the end of wednesday?\newline\# (Answer): \newline Variables = \{num\_balls\_michael\_started, num\_balls\_tuesday\_michael\_lost, num\_balls\_wednesday\_michael\_lost, num\_balls\_wednesday\_michael\_have\}\newline=> F(Rule1) => num\_balls\_michael\_started = 58\newline=> F(Rule2) => num\_balls\_tuesday\_michael\_lost = 23\newline=> F(Rule3) => num\_balls\_wednesday\_michael\_lost = 2 => total\_num\_balls\_michael\_lost = num\_balls\_tuesday\_michael\_lost + num\_balls\_wednesday\_michael\_lost = 25\newline=> F(Question) => num\_balls\_wednesday\_michael\_have = num\_balls\_michael\_started - total\_num\_balls\_michael\_lost = 58 - 25 = 33.}
       \newline-------\newline Context: \texttt{\{\{CONTEXT\}\}} 
        \\
        \hline
    \end{tabular} 
} 
\end{table*}

   

\end{document}